\documentclass[10pt,twocolumn,letterpaper]{article}

 \usepackage[pagenumbers]{cvpr} %

\makeatletter
\renewcommand\paragraph{\@startsection{paragraph}{4}{\z@}{1.0ex \@plus .3ex \@minus .2ex}{-.5em}{\normalfont\normalsize\bfseries}}
\makeatother

\usepackage{algorithm}
\usepackage{algorithmic}
\usepackage{array}

\usepackage{newfloat}
\usepackage{listings}
\usepackage{placeins}
\usepackage{booktabs} %
\usepackage{xfrac}    %

\usepackage{amsmath}
\usepackage{bm}
\usepackage{amssymb}   %

\usepackage{nameref} 
\usepackage{titletoc}

\expandafter\ifx\csname iflatexml\endcsname\relax\expandafter\newif\csname iflatexml\endcsname\fi

\newcommand{\fittable}[1]{\resizebox{\ifdim\width>\columnwidth\columnwidth\else\width\fi}{!}{#1}}

\DeclareMathOperator{\atantwo}{atan2}
\DeclareMathOperator{\mha}{MHA}

\newcommand{\timeidx}{t}     %
\newcommand{\agentidx}{g}    %
\newcommand{\stepidx}{\tau}  %

\newcommand{\query}{\bm{q}}          %
\newcommand{\key}{\bm{k}}            %
\newcommand{\dhead}{d_{\mathrm{h}}}  %
\newcommand{\hidx}{h}                %
\newcommand{\attnlogit}{\ell}        %

\newcommand{\bias}{b}                             %
\newcommand{\biasvec}{\bm{b}}                     %
\newcommand{\posmlp}{\mathrm{MLP}_{\mathrm{pos}}} %
\newcommand{\posemb}{\gamma}                      %
\newcommand{\reloff}{\bm{r}}                      %
\newcommand{\coordoff}{\bm{c}}                    %

\newcommand{\pos}{\bm{p}}                 %
\newcommand{\rotmat}{\bm{R}}              %
\newcommand{\yaw}{\psi}                   %
\newcommand{\pitch}{\vartheta}            %
\newcommand{\bearing}{\beta}              %
\newcommand{\elevation}{\alpha}           %
\newcommand{\relpos}{\Delta\pos}          %
\newcommand{\vel}{\boldsymbol{v}}         %
\newcommand{\acc}{\bm{a}}                 %
\newcommand{\yawrate}{\omega}             %
\newcommand{\heading}{\phi}   %

\newcommand{\hdim}{C}                       %

\newcommand{\rawfeat}{\bm{x}}                      %
\newcommand{\tokemb}{\bm{z}}                       %
\newcommand{\projenc}{\bm{W}_{\!\mathrm{enc}}}     %
\newcommand{\projdec}{\bm{W}_{\!\mathrm{dec}}}     %
\newcommand{\nmm}{N_{\mathrm{mm}}}

\newcommand{\dectok}{\bm{d}}                            %

\newcommand{\nuni}{N_{\mathrm{uni}}}

\newcommand{\maskspec}{\xi}                     %
\newcommand{\window}{\mathcal{T}_{\maskspec}}   %
\newcommand{\abar}{\bar{\alpha}}     %
\newcommand{\gauss}{\mathcal{N}}     %
\newcommand{\snoise}{\bm{\epsilon}}  %
\newcommand{\ident}{\bm{I}}          %

\newcommand{\dstate}{\bm{s}}                 %
\newcommand{\dstep}[1]{#1^{(\stepidx)}}      %
\newcommand{\dclean}[1]{#1^{(0)}}            %
\newcommand{\maskstate}{\textsc{mask}}          %
\newcommand{\revealp}{\rho_{\stepidx}}          %

\newcommand{\loss}{\mathcal{L}}

\usepackage{xcolor}      %

\definecolor{cvprblue}{rgb}{0.21,0.49,0.74}
\usepackage[pagebackref,breaklinks,colorlinks,allcolors=cvprblue]{hyperref}

\newcommand{\dsname}{\textup{\textsc{DIVERT}}\xspace}
\newcommand{\modelname}{\textup{\textsc{ID-MAE}}\xspace}

\def\paperID{*****}
\def\confName{CONF}
\def\confYear{YYYY}

\title{Multimodal Masked Diffusion Autoencoder for Modeling Impaired Driving}

\iflatexml
\author{John Gideon$^{*}$
  \and Anshul Gupta$^{*}$
  \and Megan Applegate-Kenton \and Kayli Battel \and Todd J Rowell
  \and Patricio Reyes Gomez \and Laporsha Dees \and Emily Sumner
  \and Avinash Balachandran \and Guy Rosman
  \thanks{$^{*}$Equal contribution. All authors are with the Toyota Research Institute, USA. Email: \texttt{firstname.lastname@tri.global}. Any opinions, findings, and conclusions expressed in this material are those of the authors and do not necessarily reflect the views of TRI or any other Toyota entity.}}
\else
\author{%
John Gideon\thanks{Equal contribution.} \quad Anshul Gupta\footnotemark[1] \quad Megan Applegate-Kenton \quad Kayli Battel \quad Todd J Rowell \\
Patricio Reyes Gomez \quad Laporsha Dees \quad Emily Sumner \quad Avinash Balachandran \quad Guy Rosman \\[4pt]
Toyota Research Institute \thanks{All authors are with the Toyota Research Institute, USA. Email: \texttt{firstname.lastname@tri.global}. Any opinions, findings, and conclusions expressed in this material are those of the authors and do not necessarily reflect the views of TRI or any other Toyota entity.}
}
\fi

\begin{document}
\maketitle
\begin{abstract}
Human cognitive state (e.g., distraction, stress), gaze, and behavior interact with the driving environment in subtle ways that AI could use to assist drivers, but ground-truth labels for human cognitive state levels are costly to obtain at scale. We propose a multimodal masked-diffusion autoencoder, \modelname, that adds a self-supervised objective to impairment prediction: alongside predicting the impairment label as a masked token, it reconstructs masked gaze and trajectory. To probe the effects of different modalities and types of impairments on model inference, we introduce \dsname, a driving-simulator dataset containing 103 hours of driving data (n = 109) with ten impairment inductions spanning cognitive, visual, and emotional states, recorded with gaze, vehicle state, and scene video. We demonstrate how \modelname affords significant improvements in impairment inference results on both the \dsname dataset and existing datasets, and enables strong results in additional tasks, such as trajectory and gaze prediction.  For location-matched pairs, our model achieves a paired AUC of 0.706 for impairment detection on \dsname, compared to 0.590 without reconstruction loss, improving over existing baselines. We probe the design space in terms of input modalities, masking scheme, positional encoding, attention mechanism, and decoding. We also investigate the interaction between various impairments in training and inference, demonstrating that visually demanding distractions are separable from cognitive distractions. We plan to publicly release the code and data.
\end{abstract}
    
\section{Introduction}
\label{sec:intro}

\begin{figure}[t]
  \centering
  \includegraphics[width=1.\columnwidth]{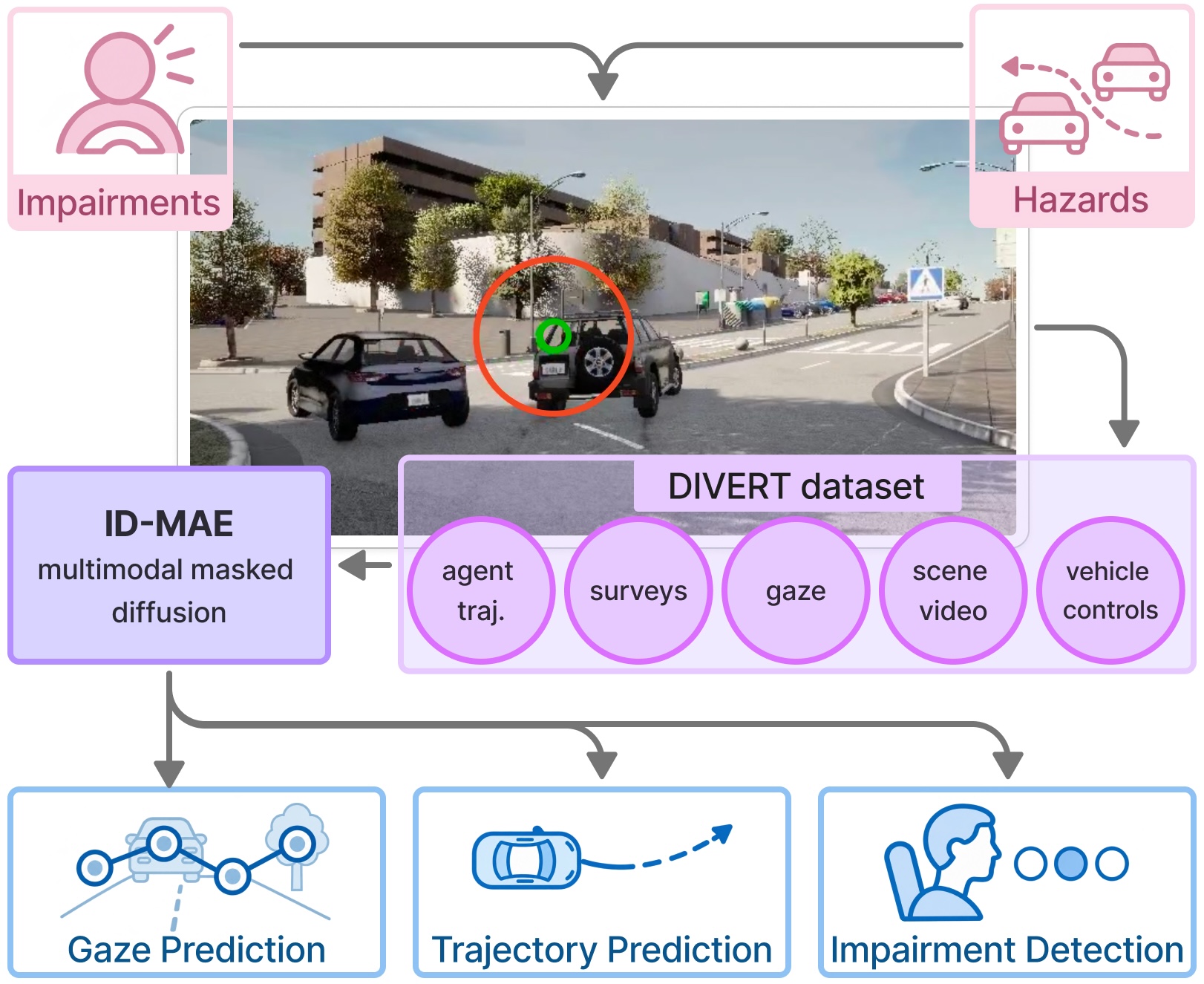}
  \caption{\modelname treats gaze (red and green circles), trajectory, scene video, and impairment as tokens of a
masked-diffusion model; the single network infers driver state when impairment
is masked and forecasts gaze and trajectory when their tokens are masked.
We trained and evaluated on \dsname, (gaze, agent trajectories, scene video under diverse impairments and
hazards), enabling detection of impairments, comparison of impaired and baseline drivers, evaluation of modality impact on impairment inference, and analysis of impairment type interactions.}
  \label{fig:teaser}
\end{figure}

The interaction of human cognitive state, gaze, and behavior lies at the core of human safety approaches~\cite{recarte2003mental,ziv2016gaze}. Reasoning about driver safety require us to understand the interplay between these over prolonged periods of time and across several modalities. For instance, consider reasoning about a driver ignoring a stopped vehicle in front of them -- their fixations on the dashboard or nearby objects may be signs of whether they are distracted, tasked, or tired. However, the space of all multimodal behavior and attention sequences is prohibitively large. Within the distribution of these sequences, risk-critical factors may not be strongly observed; for example, cognitive distraction may surface only as a slight narrowing of visual scanning and marginally delayed responses to peripheral events. Self-supervised approaches for temporal signal analysis have proven invaluable~\cite{oord2018representation, eldele2021time}, especially where labels are scarce~\cite{zhang2022self} or noisy~\cite{graf2025self}. In inherently multimodal inference tasks, such as those connecting gaze, behaviors, and latent cognitive factors, leveraging self-supervision becomes a challenging yet crucial aspect of the solution. A limited amount of data for risk-relevant, rare events~\cite{gideon2025simulator,dingus2016driver}, as well as the high-dimensional nature of the behavior observation sequences, thus requires us to draw supervision from the behavior itself rather than from labels alone.

We probe self-supervision on driving data with impairment type labels, formulating it as multimodal masked-diffusion inference, and examine how jointly modeling several impairments and driver behavior affects impairment detection. This requires richer data than existing impaired-driving datasets, which are small~\cite{shen2022cocatt}, lack scene video or agent trajectories~\cite{keshtkaran2024wacv}, are collected under different levels of vehicle automation~\cite{dargahi2024multimodal}, or limited in impairment categories~\cite{huang2025gaze}. Thus, the contributions of this work include:

\begin{itemize}
    \item We propose Impaired Driving MAE (\modelname), a model leveraging gaze, trajectories, scene video, and driver impairment state via a masked diffusion approach. It uses multimodal self-supervision and joint training on impairment tasks for improved impairment detection, while also generating gaze and trajectory predictions.
    \item We introduce Driving with Impairments in Varied Everyday Realistic Tasks (\dsname): a new dataset of diverse, controlled driver impairments, scenarios, and hazards. 
    \item We establish baselines on \dsname{} and demonstrate that \modelname{} outperforms existing baselines on \dsname{} and the Impaired Driving Dataset (IDD) \cite{gideon2025simulator}. Masked reconstruction of behavioral modalities substantially improves impairment detection over label supervision alone. We analyze per-impairment performance and report the benefit of joint training across impairment types.
\end{itemize}

\section{Related works}
\label{sec:relatedworks}
\paragraph{Driver impairment and state detection.}
Cognitive distraction has long been characterized through its effects on visual scanning, workload, and vehicle control~\cite{recarte2003mental,strayer2011cognitive},
and can be detected from behavioral and physiological signals using learned models~\cite{Liang2014_58}.
A large body of work targets automatic detection of impaired or otherwise negative driver states~\cite{seaman2025identifying}, from vehicle-control, physiological, gaze, and video signals~\cite{liang2007real,misra2023detection,zhang2026eyecue,yuan2021owipa,li2020detection,abbas2022hypo,sanikal2025driving,yang2023real}.
Multimodal models with gaze/attention have explored baseline and impaired driving (see~\cite{xing2020ensemble,akbiyik2023leveraging} and references therein), with recent datasets adding modalities~\cite{dargahi2024multimodal} or latent factors~\cite{gideon2025simulator}.
Because impairment labels are scarce, some works reduce supervision via visual contrastive learning~\cite{kopuklu2021driver,zhou2024selfsupervised} and with unsupervised learning over vehicle, physiological, and surrounding-object signals~\cite{qiu2023unsupervised}, while other works probe generalization to unseen individuals~\cite{rayatdoost2021subject,cao2022pirl,banerjee2021driver}.
We instead train one generative model whose masked targets include the impairment label alongside gaze and trajectory. Reconstructing behavior provides self-supervision, and supervising impairment types jointly can improve detection.

\paragraph{Masked modeling of multimodal sequences.}
Masked autoencoders (MAEs) have been extended to video~\cite{feichtenhofer2022masked,girdhar2023omnimae} and other modalities~\cite{bachmann2022multimae,mizrahi20234m}, unified sequence models handle many tasks and modalities as token sequences with masked denoising among their pretraining objectives~\cite{lu2022unifiedio}, and iterative unmasking~\cite{chang2022maskgit} underlies many generative variants. For motion data, masked trajectory models randomize which parts of a trajectory are hidden to support prediction, imputation, and representation learning~\cite{wu2023masked,ngiam2022scene}. Joint masking over 2D and 3D representations has also been used for point-cloud pretraining~\cite{guo2023joint}, motivating our geometry-aware masking. We combine modality-level masking with temporal windows so a single task-agnostic objective covers prediction and imputation, as well as impairment inference when the label is masked.

\begin{figure*}[t]
    \centering
    \includegraphics[width=0.95\linewidth]{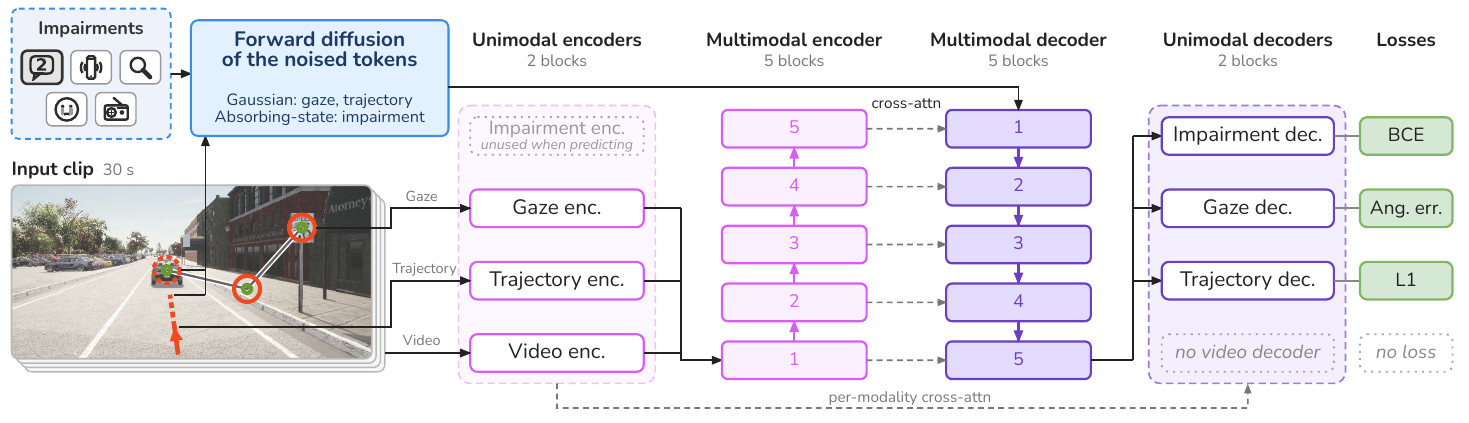}
    \caption{Architecture overview. Gaze, trajectory, video, and impairment tokens are split into \emph{clean} tokens, which the model sees as context, and \emph{noised} tokens (dashed), which it learns to reconstruct. Unimodal encoders embed each modality, and a multimodal encoder fuses them. Decoders then denoise the noised tokens in a U-Net-like fashion. Video serves as encoded context only and is not decoded.}
    \label{fig:architecture}
\end{figure*}

\paragraph{Diffusion for representation learning and multimodal sequences.}
Diffusion models have been used for self-supervised pretraining~\cite{baranchuklabel,mittal2023diffusion,xiang2023denoising} and semi-supervised learning~\cite{you2023diffusion,ho2023diffusion}. MAE diffusion has been used for representation learning~\cite{wei2023diffusion,hansen2024unified} and prediction~\cite{long2025one}, with masked generation complementing denoising-only training~\cite{wu2023denoising,tu2026masked} and encouraging information bottlenecks~\cite{wu2024macdiff}. Our decoder uses per-token noise level, similar to~\cite{chen2024diffusion}; rather than sampling independent levels, we share one sampled step across the noised tokens, tying the noise to the modality and window mask, and supply the level to the decoder as an attention bias. We jointly diffuse continuous signals (gaze, trajectory) and a discrete label using an absorbing-state process~\cite{austin2021structured}, closely related to masked discrete diffusion for language~\cite{sahoo2024simple,lou2024discrete,nie2025large} and related approaches~\cite{zheng2024masked}.
Multimodal diffusion for generation has been explored with continuous latents~\cite{bao2023one}, masked discrete tokens~\cite{swerdlow2025unified}, and mixed continuous and discrete modalities via a language-modeling loss combined with diffusion~\cite{zhou2024transfusion} or a joint process over arbitrary state spaces~\cite{rojas2025diffuse}; masked diffusion has also been used for audio-video representation learning~\cite{nunez2023diffusion}. These mostly target generation, whereas we infer a latent state and impute or predict missing behavior. Beyond driving, diffusion enables imputation of multivariate time-series~\cite{tashiro2021csdi,rasul2021timegrad,alcaraz2023diffusion}; time-series foundation models~\cite{goswami2024moment,ansari2024chronos,das2024decoder} focus on univariate series and do not leverage geometric relations between signals. Latent-variable models~\cite{johnson2016composing}, including diffusion variants~\cite{deja2023learning,kaleta2024mediffusion}, have mostly addressed isolated prediction or classification.

\paragraph{Trajectory and gaze prediction.}
Trajectory prediction has standard benchmarks~\cite{ettinger2021large,wilson2021argoverse}, strong attention-based models~\cite{ngiam2022scene,nayakanti2023wayformer,shi2022motion}, and diffusion-based forecasters~\cite{jiang2023motiondiffuser,mao2023leapfrog,zhou2026diffusiontrajpred}. These forecast agent motion from scene context alone, without driver gaze or latent state; 
Gaze modeling spans saliency~\cite{itti1998model,huang2015salicon}, scanpath prediction~\cite{assens2018pathgan,kummerer2022deepgaze}, focus of attention~\cite{palazzi2018predicting,fang2021dada}, and, recently, diffusion-based gaze generation~\cite{jiao2026diffgaze,kara2026diffeye}. Some works have used gaze and field-of-view to condition ego-trajectory prediction~\cite{akbiyik2023leveraging,liu2021avgcn}. We instead jointly model gaze, trajectory, and scene.

\section{Model}

Our model (\Cref{fig:architecture}) is an encoder--decoder transformer trained as a masked diffusion model in the style of DiffMAE~\cite{wei2023diffusion}. We extend it from images to
four modalities (gaze, trajectory, video and impairment), diffusing the impairment label as a discrete token alongside the continuous signals, and incorporate geometric relative attention.

Each modality is a grid of tokens over a time axis $T$ and a within-frame axis $W$ (agents for trajectory, image patches for video, $W{=}1$ for gaze and impairment).  Visible (\emph{clean}) tokens are embedded by per-modality \emph{unimodal encoders} and fused by a \emph{multimodal encoder}. Hidden (\emph{noised}) tokens are corrupted by forward diffusion and denoised by a \emph{multimodal decoder} followed by per-modality \emph{unimodal decoders}. Video is context only: its masked tokens are hidden from the encoder but never decoded or supervised.

\subsection{Geometric Relative Attention}
\label{sec:attention}

Every attention layer adds a learned bias~\cite{raffel2020exploring} to the scaled dot-product logit $\attnlogit$, computed from the geometry between the two tokens, similar to trajectory prediction models that encode the pairwise relative pose between agents~\cite{cui2023gorela,zhou2023query,zhang2023real}, but extended to other modalities:
\begin{equation}
    \attnlogit_{ij}^{(\hidx)} =
    \frac{\big\langle \query_i^{(\hidx)}, \key_j^{(\hidx)} \big\rangle}{\sqrt{\dhead}}
    + \bias_{ij}^{(\hidx)}, \hspace{1em}
    \biasvec_{ij} = \posmlp\!\big(\posemb(\reloff_{ij})\big),
    \label{eq:score}
\end{equation}
where $\query_i^{(\hidx)}$ and $\key_j^{(\hidx)}$ are the query and key of head $\hidx$, $\dhead$ is the per-head dimension, and $\reloff_{ij}$ collects the offsets from token $i$ to token $j$. Each offset is embedded sinusoidally ($\posemb$) and a multilayer perceptron (MLP) $\posmlp$, learned separately per query--key modality pair and layer, maps the embeddings to one bias per head.

The offsets are the differences in time index, agent index and diffusion step, plus a geometric offset. Clean tokens have step zero, so the step offset is how a noised query learns its timestep. Between two sensor tokens (gaze, or a video patch's center), the geometric offset is the difference in viewing direction $(\yaw, \pitch)$. Otherwise, it is the key's displacement, bearing and elevation as seen from a viewpoint: the query agent's heading for two trajectory tokens, or the ego looking along the sensor token's viewing direction for a sensor token and an agent. The impairment token acts as the ego looking straight ahead. Time, positions and viewing directions are never token content; they reach the model only through this bias (see details in Supplementary).

\subsection{Unimodal Encoders}
\label{sec:uni-enc}
\begin{table}[t]
    \centering
    \footnotesize
    \setlength{\tabcolsep}{3pt}
    \begin{tabular}{@{}lrrlcl@{}}
        \toprule
        Modality & Hz & $W$ & Clean token content $\rawfeat$ & Attn. & Decoded \\
        \midrule
        Impair. & 10 & 1  & one-hot ($3K$)                     & $T$    & logits ($K$) \\
        Gaze    & 10 & 1  & validity (4), events (12)          & $T$    & $\yaw, \pitch$ (4) \\
        Traj.   & 2  & 8  & $\vel, \acc, \yawrate$, flags (9)  & $T, W$ & $\acc, \yawrate$ (4) \\
        Video   & 2  & 21 & ViT-B/32 patch (768)               & $T, W$ & -- \\
        \bottomrule
    \end{tabular}
    \caption{Per-modality tokens: rate, within-frame size $W$, encoder input channels,
    attention axes in order, and decoded channels.}
    \label{tab:modalities}
\end{table}

For a token $u$ of modality $m$, the unimodal encoder linearly projects its content $\rawfeat_u$ to the model width, $\tokemb_u = \projenc^{m}\,\rawfeat_u$, and refines it with $\nuni{=}2$ transformer blocks along the modality's native axes (\Cref{tab:modalities}).
\begin{itemize}[leftmargin=*,nosep]
    \item \emph{Impairment} tokens carry the clip's labels as a one-hot over
    $\{\textsc{neg}, \textsc{pos}, \textsc{mask}\}$ for each of the $K$ impairment types. Labels are replicated at 10\,Hz across the clip to align in time with sensor tokens; all impairment tokens are masked or revealed together.
    \item \emph{Gaze} is per-eye yaw and pitch, back-projected from pre-calibrated display coordinates. The tokens carry per-eye validity flags and gaze events: saccade and fixation labels and their transitions~\cite{dalmaijer2014pygaze}. The angles enter only through the attention bias, making them relative to scene objects. 
    \item \emph{Trajectory} covers the ego vehicle and up to seven surrounding agents, translated and rotated into the ego's mean visible pose. The tokens carry velocity $\vel$ and acceleration $\acc$ in the agent's heading frame, yaw rate $\yawrate$, and validity and ego flags. Position and heading enter only through the attention bias, thus are relative to each agent's frame of reference. 
    \item \emph{Video} uses frozen vision transformer (ViT-B/32)~\cite{dosovitskiy2021an} features on a
    $3{\times}7$ grid cropped to the driving field of view.
\end{itemize}
 
\subsection{Multimodal Encoder and Decoder}
\label{sec:mm}

The multimodal encoder stacks $\nmm=5$ blocks over the unimodal encodings $\tokemb_u$ of the clean tokens. Each block has a \emph{temporal} layer, in which a token attends to the clean tokens of its own modality over time within a local window, and a \emph{cross-modal} layer, in which it attends to the clean tokens of all modalities at the nearest time index. A 10\,Hz gaze or impairment token thus reads the nearest 2\,Hz trajectory and video tokens, and the bias keeps the residual time offset (\Cref{sec:attention}). Each modality has its own query and output projections and feed-forward block, and in the cross-modal layer each query--key modality pair has its own key and value projections and its own $\posmlp$. The output of every block, $\tokemb^{(1)}, \dots, \tokemb^{(\nmm)}$, is cached.

The multimodal decoder acts on the noised tokens of the decoded modalities. A noised token carries the noised version of the clean content, with channels that would leak ground truth (validity, ego flag, gaze events) withheld:
\begin{itemize}[leftmargin=*,nosep]
    \item \emph{Impairment}: the one-hot set to \textsc{mask}.
    \item \emph{Gaze}: no content; token enters as a learned embedding.
    \item \emph{Trajectory}: the diffused acceleration $\acc$ and yaw rate $\yawrate$, and the velocity $\vel$ re-integrated from the diffused $\acc$.
\end{itemize}
The noised coordinates (gaze angles, and the trajectory position and heading integrated from
the diffused controls) are seen only through the attention bias. A per-modality linear layer
projects the noised content $\dstep{\rawfeat}_u$ to the initial decoder state
$\dectok_u = \projdec^{m}\,\dstep{\rawfeat}_u$. The decoder mirrors the encoder's layers and
parameterization and is built in reverse order: in each layer of its $n$-th block, a
noised token first cross-attends to the cached encoder output at depth $\nmm{+}1{-}n$,
\begin{equation}
    \dectok_u \leftarrow \dectok_u + \mha\big(\dectok_u,\, \tokemb^{(\nmm+1-n)}\big),
    \label{eq:dec-crossattn}
\end{equation}
with multi-head attention ($\mha$) the biased attention of Eq.~\eqref{eq:score}, reading the same clean keys as the
corresponding encoder layer, and then self-attends among the noised tokens. This U-Net-like
coupling follows the diffusion masked autoencoder of \citet{wei2023diffusion}. The diffusion
processes, the trajectory control integration, and the local attention window are detailed in the Supplementary Material.

\subsection{Unimodal Decoders}
Each decoded modality then refines $\dectok_u$ with $\nuni{=}2$ blocks along its native axes
(\Cref{tab:modalities}), mirroring its unimodal encoder. Its $n$-th block
cross-attends to the output of block $\nuni+1{-}n$ of the same modality's unimodal encoder, so the
first decoder block reads the deepest unimodal features, and then self-attends among that
modality's noised tokens. A LayerNorm and a linear head produce the prediction:
\begin{itemize}[leftmargin=*,nosep]
    \item \emph{Impairment}: one logit per impairment type at every token.
    \item \emph{Gaze}: per-eye yaw $\yaw$ and pitch $\pitch$.
    \item \emph{Trajectory}: acceleration $\acc$ and yaw rate $\yawrate$, integrated into
    velocity, position and heading (similar to~\cite{cui2020deep,salzmann2020trajectron,ivanovic2020mats}).
\end{itemize}
Following \citet{wei2023diffusion}, the trajectory and gaze heads directly predict the clean targets rather than the noise.

\subsection{Masking and Loss}
\label{sec:training}

\paragraph{Masking.}
Each training example draws a mask in three steps. First, a non-empty subset of the continuous
modalities (trajectory, gaze, video) is kept \emph{available}; the others
are dropped from the input, and the impairment label is always available. Second, a
non-empty subset of the available modalities is \emph{noised}. It must include at least one
decoded modality, and it may include the impairment label, so an example trains reconstruction with the label as context, label prediction, or both. Third, the
noised continuous modalities share one contiguous window $\window$ of length
$4$--$12$\,s, chosen with equal probability to end at the clip end (future
prediction) or to lie in the clip interior (interpolation). Tokens inside $\window$ are
noised, and the impairment label, when noised, is masked for the whole clip.

\paragraph{Loss.}
Trajectory and gaze are trained with $L_1$ losses $\mathcal{L}_{\text{traj}}$ and
$\mathcal{L}_{\text{gaze}}$ between the predicted and true values of their targets ($\acc$
and $\yawrate$ for trajectory, $\yaw$ and $\pitch$ for gaze), averaged over the noised
tokens; $\mathcal{L}_{\text{traj}}$ adds the same term computed over the ego vehicle alone.
The impairment loss $\mathcal{L}_{\text{imp}}$ is a binary cross-entropy (BCE) on the masked labels
of each impairment type, averaged over the clip's tokens.
It compares that impairment against baseline clips, with a positive-class
weight equal to the ratio of negative to positive training clips. The total loss is
\begin{equation}
    \loss = \loss_{\mathrm{traj}} + \loss_{\mathrm{gaze}} + \lambda\,\loss_{\mathrm{imp}},
    \qquad \lambda = 10,
\end{equation}
where $\loss_{\mathrm{imp}}$ averages the per impairment type losses. Further details on mask sampling,
loss weighting and implementation are in the Supplementary Material.

\section{Dataset}
\label{sec:dataset}

\begin{figure*}[t]
    \centering
    \hspace*{\stretch{1}}%
    \begin{subfigure}[t]{0.26\linewidth}
        \centering
        \includegraphics[width=\linewidth]{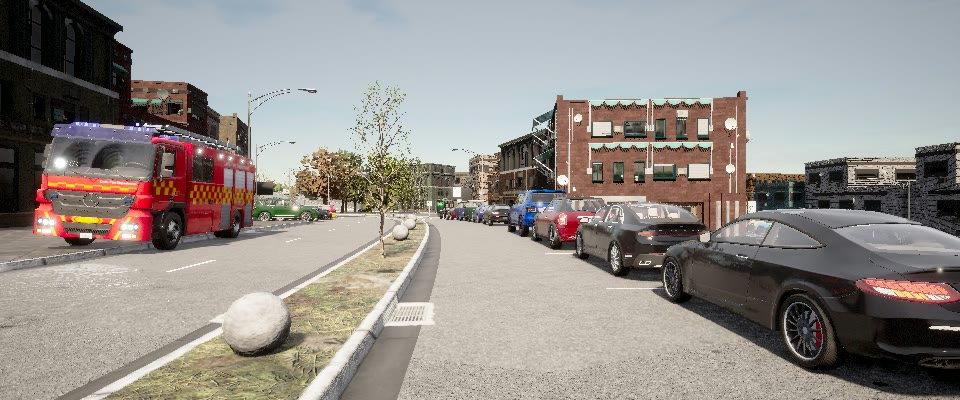}
        \caption{Visual Sink.}
    \end{subfigure}\hspace{\stretch{2}}
    \begin{subfigure}[t]{0.26\linewidth}
        \centering
        \includegraphics[width=\linewidth]{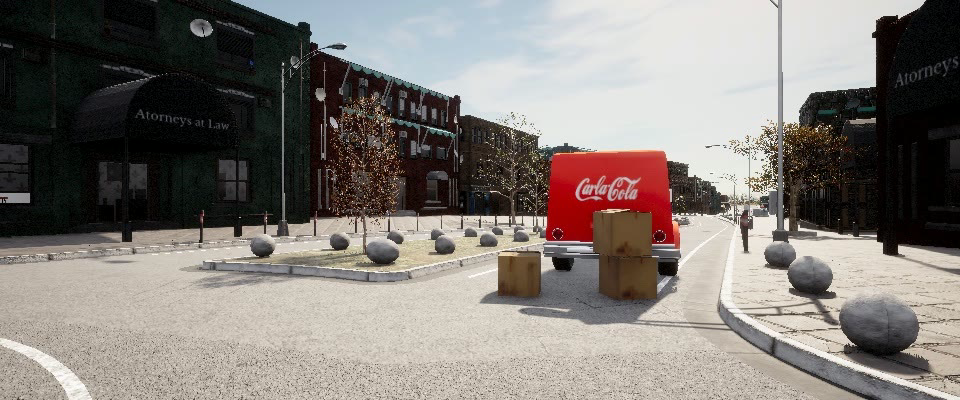}
        \caption{Forced Reroute.}
    \end{subfigure}\hspace{\stretch{2}}
    \begin{subfigure}[t]{0.26\linewidth}
        \centering
        \includegraphics[width=\linewidth]{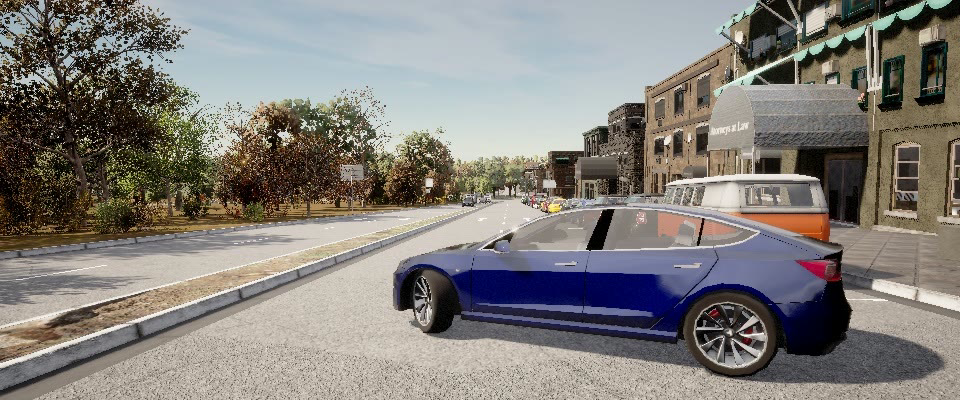}
        \caption{Hazard vehicle.}
    \end{subfigure}%
    \hspace*{\stretch{1}}
    \caption{Images of CARLA~\cite{Dosovitskiy17} environment scenes: (a) emergency vehicle props on the side of the road causing a visual sink, (b) Carla Cola props blocking the road and causing a forced reroute, and (c) the ego driver approaching a parallel parked pop-out hazard vehicle.}
    \label{fig:various_scenes}
\end{figure*}

\begin{table*}[t]
    \small
    \centering
    \begin{tabular}{c c c c c c}
        Dataset & \# Impairment Inductions (Categories) & \#Hazard Categ. & Released Modalities & Subjects & Hours \\
        \toprule

CoCAtt \cite{shen2022cocatt}     & 1 (cog.\ distraction)\textsuperscript{a} & 0 & G, S, V, L & 11 & 11.9 \\

Keshtkaran et al. \cite{keshtkaran2024wacv}       & 1 (alcohol) & $\geq$1\textsuperscript{b} & S, C & 60 & 30 \\

DATAD \cite{huang2025gaze}       & 1 (phone distraction before takeover)\textsuperscript{a} & 4 & G, S\textsuperscript{c}, V, T & 60 & n/r \\

IDD \cite{gideon2025simulator}        & 3 (alcohol, cog.\ distraction) & 4 & G, S, V, Q, T & 52 & 23.7 \\
manD 1.0 \cite{dargahi2024multimodal}       & 19 (cog. load, emot., visual/manual distraction)\textsuperscript{a, d} & 6 & G, C, V, P, Q, T & 50\textsuperscript{e} & n/r \\

C42CN \cite{taamneh2017multimodal}       & 5 (cog., emot., and sensorimotor distraction) & 1 & G, S, C, V, P, Q & 68 & $\sim$80\textsuperscript{f} \\

DMD \cite{ortega2020dmd}       & 0 (fatigue, cog., vis., man. distraction)\textsuperscript{g} & 0 (on-road) & C & 37 & 41 \\

\dsname (ours)                   & 10 (cog. load, emot., vis., cog. distraction)        & 6 & G, S, V, Q, T & 109 & 103.3 \\

         \bottomrule
    \end{tabular}
    \caption{Summary of impaired driving datasets. n/r~=~not reported. Modalities: G~=~gaze, S~=~scene video (forward, egocentric, roof, or rear view), C~=~in-cabin video, V~=~vehicle state/controls (speed, odometry, GPS, pose), P~=~physiology, Q~=~surveys, T~=~agent trajectories, L~=~annotated driver intention. \textit{
    \textsuperscript{a}Partly or wholly under vehicle automation (SAE L1--L3). 
    \textsuperscript{b}Hazards described (``such as animals entering the road'') but not enumerated. 
    \textsuperscript{c}Instance segmentation of the scene, but not RGB video. 
    \textsuperscript{d}Eight of 19 inductions are variations of n-back and auditory digit-span levels. 
    \textsuperscript{e}11 of the 50 experienced motion sickness and are released separately, leaving a test group of 39. \textsuperscript{f}Estimated from per-session durations and n = 68. 
    \textsuperscript{g}Impairments were scripted, not induced (e.g., performing fatigued actions like yawning).}}
    \label{tab:dataset-summary}
\end{table*}

\subsection{Contribution}
\paragraph{Overview.}
While there are existing public datasets with driving impairments (see \Cref{tab:dataset-summary}), \dsname is designed to cover a wide variety of impairment categories while preserving consistent rich modalities, supplying reproducible driving environments and hazards, and providing ample scale for training multimodal deep models.

\dsname contains 103.3 hours of simulator driving data collected in CARLA~\cite{Dosovitskiy17} from 109 participants. It spans 46.42 hours of driving under ten `induction' techniques covering four impairment categories -- cognitive distraction, frustration and stress, visual distraction, and cognitive load. For more information, see the Supplementary Material.

\paragraph{Data Release.}
\dsname includes synchronized forward scene video with impairment labels, vehicle trajectories, and control inputs -- with driver-facing signals including eye gaze and periodic surveys of driver cognitive load, fatigue, and emotional state. It includes a set of six controlled road hazard categories such as parked car pop-outs and red light runners. The dataset will be made publicly available.

\paragraph{Phases.}
The dataset was collected in three `Phases,' hereafter Phase 1 (Naturalistic), Phase 2 (Impaired), and Phase 3 (Hazardous). Beginning in Phase 1, participants drove through a town setting with a gamified task as `delivery drivers’ following a bespoke GPS app, with the opportunity to earn additional monetary incentive in the form of `tips,’ as motivation for repeatedly driving 5-10 minute routes broken up by surveys. Phase 2 added different impairments during driving scenarios and Phase 3 added hazards.

\paragraph{Driving Scenarios.}
Participants were assigned to one of the three collection Phases (between-subjects), and completed a series of $\sim$5-15 minute driving scenarios within that phase -- starting in Phase~2, each under a different baseline or impaired condition (within-subjects).

\paragraph{Inductions.}
Impairment inductions were informed by prior work looking at cross-domain impaired driving study methodologies~\cite{battel2026inconsistent}, and were separated into three groups: \textbf{Passive} music of the participant's choice and radio segment; \textbf{Continuous} automated phone call, mailbox scene search, radio comprehension, n-back (1-back) task, and crying baby; and \textbf{Instantaneous} texting pop-ups, visual sinks using props in-simulation (ex. \Cref{fig:various_scenes}(a); Phase 3 only), and forced GPS reroutes with blocked roads (ex. \Cref{fig:various_scenes}(b); Phase 3 only). Inductions were framed as secondary tasks where the primary task was driving while following the GPS app.

\paragraph{Hazards.}
Starting in Phase 3, we included 21 possible pre-programmed hazards of six categories throughout the map (ex. \Cref{fig:various_scenes}(c)): red light runners (4), orthogonal parked pop-out cars (5), parallel parked pop-out cars (5), reckless driver (1), cut-off turn hazards (3), and ego vehicle turn hazards (3). The addition of hazards was intended to supplement the unintentional hazards caused by traffic behavior in the other Phases and provide opportunities for interesting combinations of a hazard occurring while a driver is impaired.

\subsection{Data Collection}
\paragraph{Protocol.} After consenting, participants put on an EmotiBit~\cite{emotibit_web}, adjusted their simulator seat, and completed gaze calibration. A video introduced the GPS-guided driving task and a practice drive was completed. In Phases~2 and 3, participants additionally watched a video explaining the route timer and performance 'tips' and watched a video introducing secondary tasks. Each experimental block followed the same structure: participants drove baseline routes, watched the scenario's secondary task video (Phases~2 and 3 only), drove a GPS-guided route, and completed surveys, including a motion sickness check, shortened NASA Task Load Index (NASA-TLX)~\cite{hart1988development}, Karolinska Sleepiness Scale (KSS)~\cite{Shahid2012}, and Positive and Negative Affect Schedule (PANAS) scale~\cite{thompson2007development}. Blocks repeated until all scenarios were complete, followed by debriefing (Phases~2 and 3) and compensation.

\subsection{Dataset Statistics}
\label{sec:dataset-stats}
\paragraph{Demographics.}
Participant ages ranged from 21 to 66 years (mean of 37.7 and standard deviation of 10.7 years) for 109 completed participants across the three collection Phases (Phase 1 $n=25$, Phase 2 $n=35$, Phase 3 $n=49$).

\paragraph{Task Load.}
 To check that our impairments were psychologically noticeable, we compared how NASA-TLX scores differed by impairment type and the mean of each participant's baseline drives. We used paired $t$-tests with Holm-corrections. N-back ($+5.0$, $p < .001$, $d_z = 1.94$) and radio comprehension ($+2.5$, $p < .001$, $d_z = 0.98$) raised overall workload. Other impairments had smaller, dimension-specific effects that were not detectable in the overall score. Forced reroute increased time pressure ($+2.3$, $p = .013$), baby crying increased frustration ($+2.5$, $p = .013$), phone call \& scene search increased physical demand ($+1.3$, $p = .034$; $+1.4$, $p = .016$), and music reduced frustration ($-1.2$, $p = .020$). These results support that inductions were effective and varied in the types of demand and strength they imposed. Full analysis results are in the Supplementary Material.

\providecommand{\FloatBarrier}{\relax}
\section{Results}

\paragraph{Datasets.}
We perform experiments on \dsname{} (Section~\ref{sec:dataset}) and, for comparison, IDD~\cite{gideon2025simulator}, a driving simulator dataset of cognitive distraction and intoxication. It contains 23.7 hours of data from 52 participants, with gaze, agent trajectories, and scene video. It covers three impairments -- a statement understanding task, 1-back, and alcohol intoxication, all of which we consider as continuous impairments.

\begin{figure}[t]
    \centering
    \includegraphics[width=\linewidth]{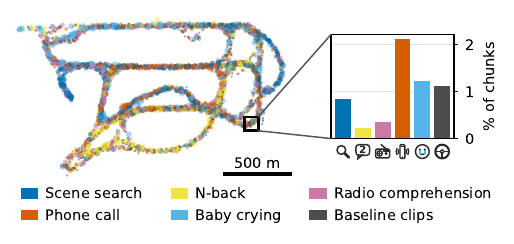}
    \caption{Spatial distribution of impairments is imbalanced so location leaks impairment data. We mitigate with paired evaluation.}
    \label{fig:paired_map}
\end{figure}

\paragraph{Labels, Clips, and Folds.}
For evaluation purposes, we consider baseline driving and passive tasks as negative samples (together considered baseline clips) and continuous impairments as positive. Instantaneous impairments are used as unlabeled data, as their label depends on where the clip falls within the scenario and when the event was triggered. We have focused on 30-second clips following IDD~\cite{gideon2025simulator}. 
\dsname is divided into five participant-disjoint folds, stratified by Phase, and IDD uses the provided folds. We select three folds as train, one as validation, and one as test. When predicting impairment, all other modalities are provided as clean input. The validation data is used to Z-normalize the logits of each binary head using baseline clips and select the maximum as a sample's overall level of impairment. We exclude \dsname practice driving and Phase~1 data from training and evaluation, as they don't contain any impairment labels.

\paragraph{Metrics.}
We then calculate the P-AUC (pooled impairment area under the curve), which represents how likely a model is to rate a continuously impaired clip higher than a baseline clip, equally weighted across impairments. We also calculate each of the ten continuous impairment pairwise combination AUCs and report the mean as I-AUC (impairment vs impairment AUC). These effectively show the model's ability to detect then differentiate types of impairment (see Supplementary for more detail on metric calculation).
All results use five round-robin runs across the folds over three seeds, with the metrics averaged across all test sets.

\newsavebox{\hlbox}
\iflatexml
  \newcolumntype{Y}{c}
  \newcolumntype{Z}{c}
\else
  \newcolumntype{Y}{>{\centering\arraybackslash}p{0.62in}}
  \newcolumntype{Z}{@{\hspace{8pt}}>{\centering\arraybackslash}p{0.62in}}
\fi
\newcommand{\dettable}{%
        \begin{tabular}{lYYZY}
            \toprule
            & \multicolumn{2}{c}{\dsname{}, matched} & \multicolumn{2}{c}{IDD~\cite{gideon2025simulator}, unmatched} \\
            \cmidrule(l{3pt}r{10pt}){2-3}\cmidrule(l{3pt}r{3pt}){4-5}
            Method & \multicolumn{1}{c}{P-AUC} & \multicolumn{1}{c}{I-AUC} & \multicolumn{1}{c}{P-AUC} & \multicolumn{1}{c}{I-AUC} \\
            \midrule
            \modelname & \textbf{0.706} & \textbf{0.728} & \textbf{0.744} & \textbf{0.691} \\
            \hspace{1em}No reconstruction & 0.590 & 0.603 & 0.689 & 0.635 \\
            \hline
            Logistic reg.~\cite{gideon2025simulator} & 0.571 & 0.624 & 0.695 & 0.631 \\
            Random forest~\cite{gideon2025simulator} & 0.525 & 0.638 & 0.681 & 0.605 \\
            Gradient boosting~\cite{gideon2025simulator} & 0.539 & 0.664 & 0.671 & 0.636 \\
            SVM~\cite{gideon2025simulator} & 0.548 & 0.685 & 0.670 & 0.663 \\
            Multi-task MLP~\cite{gideon2025simulator} & 0.571 & 0.657 & 0.700 & 0.672 \\
            \hline
            Location only & 0.529 & 0.527 & 0.502 & 0.509 \\
            \bottomrule
        \end{tabular}}
\begin{table*}[ht]
    \centering
    \begin{subtable}[c]{0.60\textwidth}
        \centering
        \small
        \setlength{\tabcolsep}{3.5pt}
        \iflatexml
            \dettable
        \else
            \global\setbox\hlbox\hbox{\dettable}\usebox{\hlbox}
        \fi
        \caption{Detection AUC.}
        \label{tab:impairment_detection_main}
    \end{subtable}\hfill
    \begin{subtable}[c]{0.385\textwidth}
        \centering
        \iflatexml
            \includegraphics[width=0.38\textwidth]{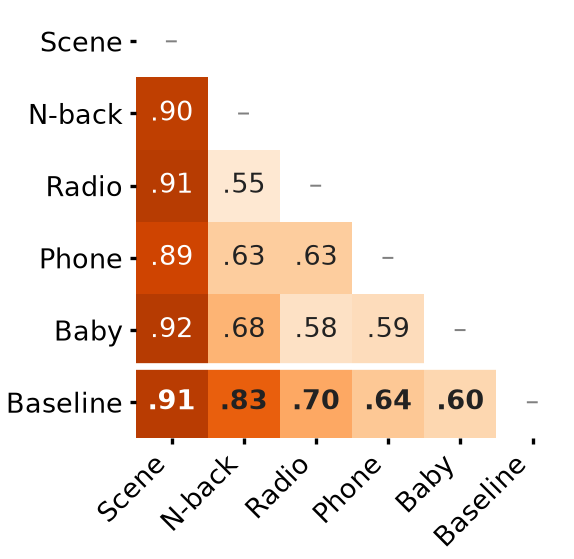}
        \else
            \includegraphics[width=\linewidth,height=\dimexpr\ht\hlbox+\dp\hlbox\relax,keepaspectratio]{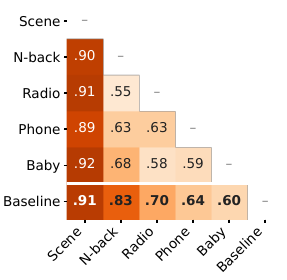}
        \fi
        \caption{\modelname paired AUCs.}
        \label{fig:confusion}
    \end{subtable}
    \caption{Impairment detection on \dsname{} and IDD~\cite{gideon2025simulator}. (a) location-matched pairs for \dsname{} and on all pairs, unmatched, for IDD. \modelname, No recon.\ and the MLP are averaged over three seeds; the other baselines are a single run. Bold: best per column. (b) Pairwise location-matched AUCs of \modelname{} on \dsname{}, with baseline clips (``Baseline'') as an additional class placed last.}
    \label{tab:impairment_detection}
\end{table*}

\paragraph{Paired Location Evaluation.}
Because \dsname crosses many impairment types with different routes and each participant drives only a small subset of the combinations, the joint distribution of impairment and route is not uniform. Initial investigations of \dsname found that a clip's location and path leaked impairment information, due to this imbalance, as shown in Figure~\ref{fig:paired_map}. A logistic regression model trained with coarse location information alone demonstrates this leak with a P-AUC of 0.602 and I-AUC of 0.810. To correct for this, we focus the evaluation on pairs of samples that start near the same location and follow a similar path (see the Supplementary Material for details). We also only sample pairs within Phase to avoid any cohort-level shortcut. When instead using the same coarse location model over these pairs, the performance decreases to a P-AUC of 0.529 and an I-AUC of 0.527, controlling for location leak in the dataset. These 30{,}929 pairs are released with the dataset. On IDD we use all pairs, as location does not predict impairment (P-AUC of $0.502$ and I-AUC of $0.509$).

\subsection{Impairment Detection}
\label{sec:impairment}

Table~\ref{tab:impairment_detection}\subref{tab:impairment_detection_main} has impairment detection results. Table~\ref{tab:impairment_detection}\subref{fig:confusion} has per-impairment AUCs vs\ baseline clips \& impairments.

\paragraph{Effect of self-supervision.}
Removing the reconstruction loss and training on the impairment heads alone lowers P-AUC from $0.706$ to $0.590$ and I-AUC from $0.728$ to $0.603$ on \dsname{} ($\Delta$ $-0.116$ and $-0.125$), and from $0.744$ to $0.689$ and $0.691$ to $0.635$ on IDD ($\Delta$ $-0.055$ and $-0.056$). It is lower in every test fold on both datasets, by $0.08$--$0.20$ per fold on \dsname{}. Masked reconstruction of the behavioral modalities therefore supplies valuable training signal that label supervision alone does not.

\paragraph{Comparison with feature baselines.}
We retrain the publicly released baselines of \cite{gideon2025simulator} on our impairment heads, negatives and folds. \modelname{} leads every baseline on every metric. The gap is largest on \dsname{}, where the baselines reach only $0.525$--$0.571$ P-AUC ($\Delta$ $-0.135$ to $-0.181$), close to location only ($0.529$), and $0.624$--$0.685$ I-AUC ($\Delta$ $-0.043$ to $-0.104$). Built on compact hand-crafted features, they remain strong in low-data regimes. On IDD, almost a third the size of leveraged \dsname{} data ($23.7$ vs.\ $65.1$ hours), they come within $0.044$--$0.074$ of \modelname{} on P-AUC ($0.670$--$0.700$) and $0.019$--$0.086$ on I-AUC ($0.605$--$0.672$), with the multi-task MLP closest overall and support vector machine (SVM) close on I-AUC.

\paragraph{Which impairments are confused.}
 The impairments form \emph{two groups}: scene search separates from every other impairment ($0.89$--$0.92$), whereas n-back, phone call, baby crying and radio comprehension are hard to tell apart ($0.55$--$0.63$; the exception, n-back vs.\ baby crying at $0.68$, rests on only 314 pairs). The groups follow the tasks' demands. Scene search is a \emph{visual} task that pulls the eyes off the road, whereas the other four are \emph{cognitive}, loading attention without a comparable visual demand. Against baseline clips, scene search ($0.91$) is most clearly detected, with  n-back ($0.83$) and radio comprehension ($0.70$) having the highest scores among cognitive tasks. Phone call ($0.64$) and baby crying ($0.60$) are only somewhat more distinguishable from baseline clips than from each other.

\subsection{Trajectory and Gaze Reconstruction}
Besides detecting impairment, \modelname{} reconstructs the masked portions of gaze and trajectory. We mask the last $8$\,s of every vehicle's trajectory and the driver's gaze, leaving $22$\,s of context, so the model forecasts $8$\,s ahead, the horizon of the Waymo Open Motion benchmark~\cite{ettinger2021large}. Tables~\ref{tab:traj} and~\ref{tab:gaze} report the ego vehicle's average / final displacement error (ADE and FDE, in meters, over the full $8$\,s) and gaze angular error (degrees, over the first $1$\,s and $2$\,s). Because the decoder is a diffusion model, we score both a \emph{single rollout} and the \emph{best of $N$} (subscript \emph{best}; the per-clip minimum, $N{=}6$ as in Waymo); the gap between the two shows how varied its sampled futures are. We evaluate:
\begin{itemize}
  \item \textbf{\modelname{}} with video visible only for the past or for the full clip, plus a \emph{single-step} (non-diffusion) variant;
  \item \textbf{Heuristic predictors}: constant velocity,  training-mean rate (trajectory); context mean,  training mean (gaze).
\end{itemize}
Qualitative results are in the Supplementary Material.

\paragraph{Trajectory.}
Every model is far better than the heuristic predictors (constant velocity ADE $10.37$\,m, FDE $26.80$\,m). Seeing the full video narrows possible futures and helps both decoders, but diffusion gains more (ADE $7.61 \to 5.11$\,m vs.\ $7.05 \to 5.28$\,m). Using the \emph{best of six} puts diffusion clearly ahead. With past video, its best of six error ($5.80$\,m ADE, $15.17$\,m FDE) is $24$--$28\%$ below its single sample and $13$--$15\%$ below single-step's best of six. Single-step's samples vary only due to input noise, so its best of six gains just $5$--$6\%$. With full video, diffusion's gain shrinks to $7$--$9\%$, as fewer futures remain plausible. Best of six is optimistic, but its gain over a single sample shows that diffusion generates diverse rollouts, some close to the ground truth.

\begin{table}[t]
    \centering
    \small
    \setlength{\tabcolsep}{3pt}
    \fittable{        \begin{tabular}{l r r r r}
        \toprule
        Model & ADE & ADE$_{best}$ & FDE & FDE$_{best}$\\
        \midrule
        Constant velocity & 10.37 & - & 26.80 & -\\
        Train mean rate & 10.51 & - & 27.20 & -\\
        \midrule
        Diffusion, past vid. & 7.61 & \textbf{5.80} & 20.99 & \textbf{15.17}\\
        Single step, past vid. & \textbf{7.05} & 6.68 & \textbf{19.09} & 17.87\\
        \midrule
        Diffusion, full vid. & \textbf{5.11} & \textbf{4.73} & \textbf{12.79} & \textbf{11.59}\\
        Single step, full vid. & 5.28 & 5.12 & 13.27 & 12.76\\
        \bottomrule
    \end{tabular}}
    \caption{Trajectory reconstruction error (m). ADE and FDE use a single sample; ADE$_{best}$ and FDE$_{best}$ use the best of six. Past vid.\ sees video only up to the prediction start; full vid.\ sees the future.}
    \label{tab:traj}
\end{table}

\begin{table}[t]
    \centering
    \small
    \setlength{\tabcolsep}{3pt}
    \begin{tabular}{l r r r r}
        \toprule
        & \multicolumn{2}{c}{Angular error @ 1\,s} & \multicolumn{2}{c}{Angular error @ 2\,s} \\
        \cmidrule(l{3pt}r{10pt}){2-3}\cmidrule(l{3pt}r{3pt}){4-5}
        Model & Single$\downarrow$ & Best$\downarrow$ & Single$\downarrow$ & Best$\downarrow$\\
        \midrule
        Visible context mean & 11.17 & - & 11.02 & - \\
        Train mean & 11.36 & - & 11.14 & - \\
        \midrule
        Diffusion, past vid. & 10.60 & \textbf{8.16} & 10.54 & \textbf{8.45} \\
        Single step, past vid. & \textbf{9.31} & 9.15 & \textbf{9.23} & 9.09 \\
        \midrule
        Diffusion, full vid. & 9.84 & \textbf{8.07} & 9.54 & \textbf{8.11} \\
        Single step, full vid. & \textbf{8.59} & 8.53 & \textbf{8.38} & 8.33 \\
        \bottomrule
    \end{tabular}
    \caption{Gaze angular error (degrees), over the first 1\,s and 2\,s of the masked window. Single uses one sample; Best the best of six.}
    \label{tab:gaze}
\end{table}

\paragraph{Gaze.}
Gaze shifts within a fraction of a second, so it is much harder to forecast than the trajectory, and prior work predicts it only for a short horizon. Driver attention models predict the current or next frame~\cite{palazzi2018predicting,xia2018predicting,fang2021dada}, and few works anticipate gaze seconds ahead (\eg 2s in~\cite{lai2024listen}). We therefore score the first $1$\,s and $2$\,s of the masked window (Table~\ref{tab:gaze}). Even there, a single diffusion sample with past video only slightly beats the visible-context mean ($10.54^\circ$ vs.\ $11.02^\circ$ at $2$\,s) and trails single-step ($9.23^\circ$). Full video helps both decoders ($9.54^\circ$ and $8.38^\circ$). The \emph{best of six} again puts diffusion ahead ($8.45^\circ$ and $8.11^\circ$ vs.\ $9.09^\circ$ and $8.33^\circ$). It lowers diffusion's error by $20\%$ with past video and $15\%$ with full video, against at most $2\%$ for single-step.

\subsection{Model Ablations and Comparisons}
\label{sec:ablation}

\begin{table}[ht]
    \centering
    \small
    \begin{subtable}[c]{0.50\columnwidth}
        \centering
        \setlength{\tabcolsep}{2pt}
        \begin{tabular}{l r r}
        \toprule
        Config. & P-AUC & I-AUC \\
        \midrule
        \modelname & \textbf{0.706} & 0.728  \\
        Only Gaze & 0.665 & 0.720  \\
        No Gaze & 0.551 & 0.544  \\
        Gaze @ 2\,Hz & 0.658 & 0.684  \\
        \midrule
        Global pos. & 0.688 & 0.706  \\
        Single step & 0.688 & \textbf{0.742}  \\
        Decode video & 0.695 & 0.729  \\
        Specialists & 0.687 & 0.731  \\
        \bottomrule
        \end{tabular}
        \caption{Ablations.}
        \label{tab:ablation_main}
    \end{subtable}\hfill
    \begin{subtable}[c]{0.47\columnwidth}
        \centering
        \iflatexml
            \includegraphics[width=0.45\textwidth]{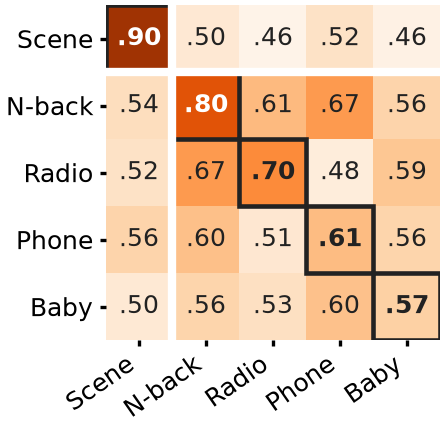}
        \else
            \includegraphics[width=\linewidth]{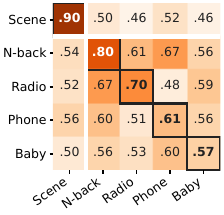}
        \fi
        \caption{Specialist transfer.}
        \label{fig:specialist_transfer}
    \end{subtable}
    \caption{(a) We perform ablations by changing one aspect of \modelname per row. Dropping gaze modality or frame rate, relative positions, or joint training decreases performance, while the other changes have mixed impact. (b) We expand on the per-impairment specialist results -- training on one and testing on another.}
    \label{tab:ablation}
\end{table}

We perform several ablations, highlighted in Table~\ref{tab:ablation}\subref{tab:ablation_main} (full set in Supplementary). We first examine the impact of the gaze modality, including only using it, dropping it, and reducing its frame rate to 2\,Hz to match the video and trajectory modalities. Both general impairment detection and differentiation rely strongly on gaze, with a gaze-only model reaching $0.665$ P-AUC and $0.720$ I-AUC. With trajectory and video added, the I-AUC stays similar, potentially because the strongest impairment differentiation scores come from scene search, a task which gaze is particularly suited to capturing. However, the P-AUC increases substantially. Notably, a model with gaze dropped has near chance performance. Reducing gaze frame rate substantially drops performance, demonstrating the benefit of a flexible model allowing different frame rates for each modality.

The ablations also show a slight benefit of geometric relative position injection over standard global positions. We find that training only with single step prediction results in similar performance to diffusion. While useful for generating trajectory and gaze rollouts, the impact to impairment prediction alone is mixed. Decoding the video tokens does not appear beneficial and increases compute.

We also trained specialist models that each only see one type of impairment and baseline to determine the impact of joint task training. Combining them together during evaluation lowers the P-AUC to $0.687$, as the trained heads are not calibrated on one another. Specialists retain a similar I-AUC because it is dominated by the scene-search vs. cognitive pairs, which each specialist separates on its own. Table~\ref{tab:ablation}\subref{fig:specialist_transfer} shows how much each impairment's specialist detects the others -- showing some transfer between cognitive impairments, but no transfer for scene search. 

\section{Conclusion}
We presented \modelname, a multimodal masked diffusion autoencoder that jointly models impairment, gaze, trajectory, and video, and \dsname, a driving-simulator dataset of 109 drivers under ten impairment inductions.
We demonstrated how subtle, data-limited tasks such as impairment detection can leverage supervision from adjacent tasks, improving paired detection from $0.590$ to $0.706$ AUC, while also enabling gaze and trajectory prediction.
The impairments and hazards in \dsname open several directions we are eager to explore: detecting impairment at lower latency, analyzing how impairment impacts behavior under hazards, and investigating the role of language through further annotations and modeling.

\section{LLM Usage Statement}

General-purpose commercial LLM assistants (Claude, Anthropic) were used in the preparation of this work. Their use is described below.

\paragraph{Writing and editing.} Copyediting and grammar checking of human-written text; condensing and restructuring passages for length; prose and figure initial drafting; and redundancy checking across the paper.

\paragraph{Related work and references.} Searching for related literature and datasets; and initial reviews of potentially relevant literature. Every dataset, reported statistic, cited literature, and bibliographic entry appearing in the paper was verified by the authors against the primary source.

\paragraph{Code.} LLM assistants were used to initially generate code, including drafting functions from author-written specifications and debugging functionality. All code was reviewed, tested, and validated by the authors, who are responsible for the correctness of the reported results.

\paragraph{Research and Collection.} No LLM was used for research goal creation or for the design or execution of the data collection. The authors take full responsibility for all content, including all scientific claims and citations, and confirm that every section of the paper was reviewed, vetted, and edited by human authors.

{
    \small
    \bibliographystyle{ieeenat_fullname}
    \bibliography{main}
}

\clearpage
\renewcommand{\dblfloatpagefraction}{0.8}  %
\setcounter{page}{1}
\maketitlesupplementary
\label{sec:Supplementary}

\startcontents[supp]
\titlecontents{section}[0pt]{\addvspace{4pt}\bfseries}{\thecontentslabel\hspace{1em}}{}{\hfill\contentspage}
\titlecontents{subsection}[1.5em]{}{\thecontentslabel\hspace{1em}}{}{\titlerule*[0.6pc]{.}\contentspage}
\printcontents[supp]{}{1}{\noindent{\large\bfseries Contents}\par\medskip\setcounter{tocdepth}{2}}

\section{Model Details}

\subsection{Geometric Offsets}
\label{supp:geometry}

\paragraph{Offset vector.}
For query token $i$ and key token $j$, the offsets of Eq.~\eqref{eq:score} are:
\begin{equation}
    \reloff_{ij} = \big(\timeidx_j - \timeidx_i,\; \agentidx_j - \agentidx_i,\;
    \stepidx_j - \stepidx_i,\; \coordoff_{ij}\big),
\end{equation}
where $\timeidx$ is the time index, $\agentidx$ the agent index (present only when both
tokens are trajectory tokens), $\stepidx$ the diffusion step, and
$\coordoff_{ij}$ the geometric offset defined below. Time indices are on a 10\,Hz grid
shared by all modalities, so tokens of a 2\,Hz modality sit at every fifth index.

\paragraph{Embedding.}
The $D$ scalar offsets of a query--key pair split the model width $\hdim{=}64$ equally,
$C' = 2\lfloor \hdim / 2D \rfloor$ channels each. Each offset is mapped to $C'/2$ sin/cos
pairs at geometrically spaced frequencies (following~\cite{vaswani2017attention}).
Angular offsets $\theta$ instead repeat $(\sin\theta, \cos\theta)$ $C'/2$ times, making
their embedding $2\pi$-periodic. The embeddings are concatenated, zero-padded to $\hdim$, and mapped by
$\posmlp$ to one bias per head.
Its last layer is zero-initialized, so training starts from unbiased attention.

\paragraph{Sensor--sensor.}
Gaze tokens and video patches have viewing directions $(\yaw, \pitch)$ relative to the screen
center (\Cref{supp:inputs}). Both look out from the ego, so they relate only through direction.
The geometric offset is the difference in viewing direction,
$\coordoff_{ij} = (\yaw_j - \yaw_i,\; \pitch_j - \pitch_i)$, computed per eye for gaze. 

\paragraph{Trajectory--trajectory.}
Let $\pos_i \in \mathbb{R}^3$ and $\heading_i$ be the position and heading of trajectory
token $i$ in the common frame (the ego's mean visible pose), and $\rotmat(\heading)$ the
rotation to heading $\heading$. The key's displacement from the query,
$\pos_j - \pos_i$, is rotated by $\rotmat(\heading_i)^{\!\top}$ into the query's
point-of-view (POV) frame, so the offset is independent of the query's absolute heading.
The key's relative position, bearing and elevation are:
\begin{equation}
    \begin{aligned}
        \relpos_{ij} &= \rotmat(\heading_i)^{\!\top}(\pos_j - \pos_i), \\
        \bearing_{ij} &= \atantwo\!\big(\relpos_{ij,y},\, \relpos_{ij,x}\big), \\
        \elevation_{ij} &= \atantwo\!\big(\relpos_{ij,z},\, \lVert \relpos_{ij,xy} \rVert\big),
    \end{aligned}
    \label{eq:pov}
\end{equation}
and the geometric offset is $\coordoff_{ij} = (\relpos_{ij}, \bearing_{ij}, \elevation_{ij})$.

\paragraph{Sensor--trajectory.}
For a sensor token $s$ and a trajectory agent $j$ (either may be the query), the POV is the
ego looking along the sensor's viewing direction $(\yaw_s, \pitch_s)$ (averaged over the eyes
for gaze) so the agent is placed relative to where the driver looks or the patch points:
\begin{equation}
    \relpos_{sj} = \rotmat_{\pitch}(\pitch_s)^{\!\top}\,
    \rotmat(\heading_{\mathrm{ego}} - \yaw_s)^{\!\top}(\pos_j - \pos_{\mathrm{ego}}),
\end{equation}
with the ego pose taken at the nearest trajectory time index and $\rotmat_{\pitch}$ tilting
the forward axis to the viewing pitch. Bearing and elevation follow as in
Eq.~\eqref{eq:pov}.

\paragraph{Impairment--any.}
The impairment token has no position or viewing direction of its own and acts as the ego
looking straight ahead, $(\yaw, \pitch) = (0, 0)$. Its offset to a gaze or video token is
that token's viewing direction, as in the sensor--sensor case. Its offset to a
trajectory agent is the agent's position in the ego frame, as in the sensor--trajectory case.

\paragraph{Noised states.}
For a noised token, the bias is computed from the noised state: the diffused gaze angles,
and for trajectory the position and heading integrated from the diffused acceleration and
yaw rate (\Cref{supp:integration}). These states are never token content, so the bias
is the only path by which the decoder sees them.

\subsection{Per-Modality Inputs}
\label{supp:inputs}

\paragraph{Gaze.}
A point on the display has normalized coordinates $(x, y) \in [0, 1]^2$, measured from the
top-left corner. Its viewing direction relative to the screen center is
\begin{equation}
    \yaw = \big(x - \tfrac12\big)\,\mathrm{FoV}_h, \qquad
    \pitch = -\big(y - \tfrac12\big)\,\mathrm{FoV}_v,
    \label{eq:display}
\end{equation}
where $\mathrm{FoV}_h$, $\mathrm{FoV}_v$ are the display's horizontal and vertical fields
of view. Each eye's pre-calibrated gaze point on the display is back-projected this way and
subsampled to 10\,Hz. 

The token content is 8 channels per eye (16 in total): validity flags
for yaw and pitch, and gaze events from PyGaze~\cite{dalmaijer2014pygaze}, namely saccade and
fixation labels, their transitions between consecutive tokens, and a validity flag for each of
labels and transitions. A transition is zeroed, and its flag cleared, when either token is
masked, so a clean token cannot encode a masked neighbor. All event channels are withheld from
noised tokens.

\paragraph{Trajectory.}
Up to eight agents (the ego vehicle and seven others, chosen by closest proximity to ego) are sampled at 2\,Hz and placed in a
common frame by translating and rotating them by the ego's mean position and heading over its
visible tokens. The token content is 9 channels: velocity $\vel$ (longitudinal, lateral and
vertical), the finite difference of position rotated into the agent's heading frame;
acceleration $\acc$, the finite difference of $\vel$; yaw rate $\yawrate$, the wrapped finite
difference of heading; a validity flag; and a flag marking whether the agent is the ego
vehicle. Position and heading are not content and
enter only through the attention bias (\Cref{supp:geometry}).

\paragraph{Video.}
Front-camera frames are encoded offline at 224$\times$224 by a frozen ViT-B/32 pretrained on
ImageNet-21k~\cite{dosovitskiy2021an}. Of its $7{\times}7$ grid of 768-d patch embeddings, the
middle three rows, which cover the image content, are kept, giving a $3{\times}7$ grid at
2\,Hz. Each patch is assigned the viewing direction of its center by Eq.~\eqref{eq:display}.

\paragraph{Impairment.}
\dsname has five impairment types (n-back, automated phone call, scene search, crying baby, radio
comprehension) and IDD three (n-back, statement task, intoxication). Each type's label is a
one-hot vector over $\{\textsc{neg}, \textsc{pos}, \textsc{mask}\}$.

\subsection{Diffusion and Masking}
\label{supp:diffusion}

\paragraph{Gaussian diffusion (gaze, trajectory).}
States are standardized with population statistics. For a step
$\stepidx \sim \mathcal{U}\{1, \dots, 1000\}$, shared by all modalities
of an example, the clean values $\dclean{\dstate}$ of the noised targets are corrupted as
\begin{equation}
    \dstep{\dstate} = \sqrt{\abar_\stepidx}\, \dclean{\dstate} + \sqrt{1 - \abar_\stepidx}\, \snoise,
    \qquad \snoise \sim \gauss(\bm{0}, \ident),
\end{equation}
with a cosine schedule~\cite{nichol2021improved} rescaled to zero terminal
SNR~\cite{lin2024common}, so that $\dstate^{(1000)}$ is pure noise. The targets are
acceleration and yaw rate for trajectory, and per-eye yaw and pitch for gaze (one noise sample
shared by the two eyes). The other noised trajectory states are recomputed from the noised
targets rather than noised themselves (\cref{supp:integration}).

\paragraph{Absorbing-state diffusion (impairment).}
Following D3PM~\cite{austin2021structured}, a noised impairment label is replaced by
\maskstate{} with probability $1 - \abar_\stepidx$ under the same schedule, with one decision
per example covering all of its tokens and impairment types. In the reverse process, a masked
label is revealed between steps $\stepidx$ and $\stepidx' < \stepidx$ with probability
$\revealp = (\abar_{\stepidx'} - \abar_\stepidx)/(1 - \abar_\stepidx)$. Each type is revealed to
a class sampled from its token-averaged logit and then stays fixed.

\paragraph{Mask sampling.}
All pairs of an available set and a noised set that satisfy the constraints of
\cref{sec:training} are enumerated, giving 41 pairs. Examples draw them without replacement
from a shuffled deck that is reshuffled when empty, so each pass through the deck covers every
pair once. 

\paragraph{Loss details.}
The $L_1$ loss is computed in degrees for the gaze angles, in degrees per second for yaw
rate, and for acceleration in standardized units: each axis (longitudinal, lateral, vertical)
Z-scored with population statistics (subtract the mean and divide the standard deviation). The positive-class weight of each impairment type is
computed from the fold's training split.

\paragraph{Sampling.}
At test time we sample with 50 deterministic DDIM steps~\cite{song2021denoising}, evenly spaced
from step 1000 to 1. The impairment label follows the absorbing-state reverse process on the
same steps. Because the encoder sees only clean tokens and receives no timestep conditioning,
its outputs are computed once and reused across all steps.

\subsection{Trajectory Control Integration}
\label{supp:integration}
Velocity, position and heading are integrated from acceleration and yaw rate, both for the
noised inputs (from the diffused controls) and for the outputs (from the predicted ones). Models that predict controls for future prediction integrate them forward from the last
observation~\cite{cui2020deep,salzmann2020trajectron,ivanovic2020mats}. Since our masks include
interior intervals, we anchor the integration at both ends. For a quantity $y$ with rate
$\dot y$, integration over a masked interval runs forward from the last clean token and
backward from the next one,
\begin{equation}
    y^{\mathrm{f}}_{\timeidx} = y^{\mathrm{f}}_{\timeidx-1} + \Delta t\,\dot y_{\timeidx-1}, \qquad
    y^{\mathrm{b}}_{\timeidx} = y^{\mathrm{b}}_{\timeidx+1} - \Delta t\,\dot y_{\timeidx},
\end{equation}
with $\Delta t = 0.5$\,s, and the two are blended by distance,
\begin{equation}
    y_{\timeidx} = \frac{d^{\mathrm{b}}_{\timeidx}\, y^{\mathrm{f}}_{\timeidx}
    + d^{\mathrm{f}}_{\timeidx}\, y^{\mathrm{b}}_{\timeidx}}
    {d^{\mathrm{f}}_{\timeidx} + d^{\mathrm{b}}_{\timeidx}},
\end{equation}
where $d^{\mathrm{f}}_{\timeidx}$ and $d^{\mathrm{b}}_{\timeidx}$ count the steps to the
previous and next clean token. For future prediction, where no clean token follows, only the
forward integration is used. 

\subsection{Local Temporal Attention}
\label{supp:local}
Temporal layers attend densely to keys within 6\,s of the query and sparsely beyond, to keys at
multiples of 6\,s. For our 30\,s clips, a token thus reads every key within 6\,s and the keys
6, 12, 18 and 24\,s away. Cross-modal layers read the nearest time index and are unaffected.
We also tried global temporal attention which performed similarly at a much higher compute cost.

\subsection{Implementation Details}
\label{supp:impl}

\paragraph{Architecture.}
All modalities use width $\hdim = 64$ and use pre-norm transformer blocks with 8 heads, an
MLP ratio of 2 and dropout $0.1$, and LayerNorms without affine parameters. The unimodal
stacks have two blocks and the multimodal stacks five, in both the encoder and the decoder.

\paragraph{Optimization.}
We train for 100 epochs with AdamW (weight decay $0.01$), a batch size of 48 and a learning
rate of $2.4\times10^{-3}$ that decays by a factor of $0.975$ per epoch, with no warmup. We use
gradient checkpointing and full precision, and evaluate the weights of the last epoch.

\section{Dataset}
\subsection{Ethics and IRB Compliance}

All data collection was conducted under WCG IRB (Protocol \#20241945), in accordance with the Declaration of Helsinki. All subjects provided written informed consent prior to participation and were compensated for their time.

\subsection{Simulator Setup}
\begin{figure}[t]
    \centering
    \includegraphics[width=\columnwidth]{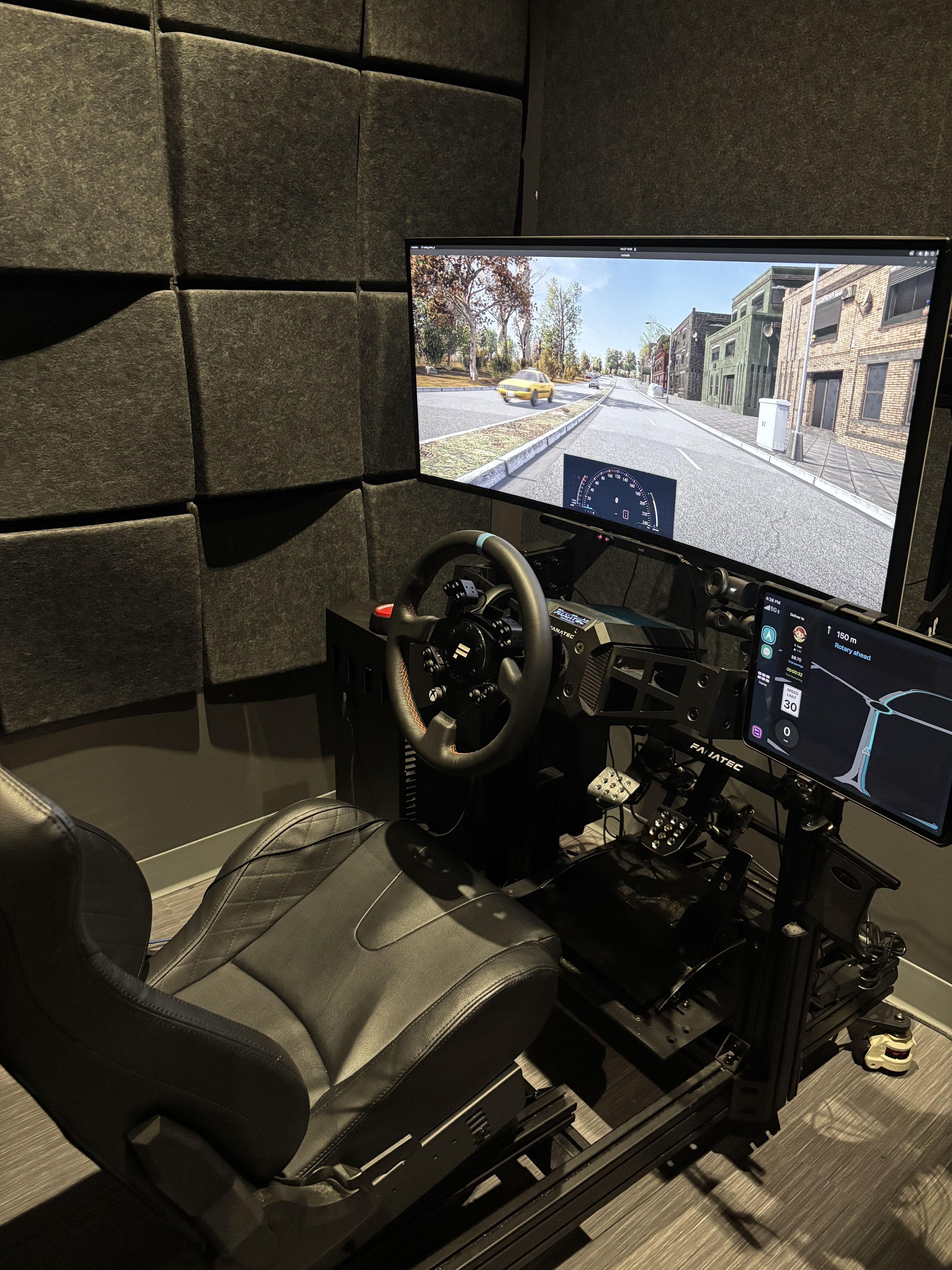}
    \caption{Image of the full fixed-base simulator seat, wheel, pedals, CARLA scene, GPS app, and red push button.}
    \label{fig:simulator_setup}
\end{figure}

The simulator rig was built on a Trak Racer TR80S frame, fitted with a Fanatec Podium DD2 direct-drive wheel base, a ClubSport R330 V2 steering wheel, and ClubSport V3 inverted pedals on a sliding mount. Participants viewed the scene on a 38-inch LG UltraWide curved display (LG 38WR85QC). All scenarios were built in CARLA~\cite{Dosovitskiy17}, with a custom scripting layer handling study flow and scenario logic. CARLA ran on a Puget Systems workstation (Intel Core i9, 128 GB RAM, 2× NVIDIA RTX 6000 Ada). Participants received navigation instructions from a map displayed on an iPad Pro (6th generation, 12.9-inch) mounted to the right of the screen, and used a separate push button to move between study phases. The sensors facing the driver were a Tobii Pro Spark eye tracker (collecting timestamp, gaze origin, gaze point, and pupil diameter), a Logitech MX Brio 4K webcam (model V-U0040), and a headset microphone. Drivers were seated about 65 cm from the display (subject to seat adjustment) with the Logitech webcam mounted to the seat above their head facing toward the screen. 
Participants wore an EmotiBit wristband (physiological and movement tracking)~\cite{emotibit_web} on the inside of one wrist. ROS~\cite{macenski2022robot} logged driver inputs, vehicle state and camera frames, and study logic telemetry. For images of the simulator setup, see~\Cref{fig:simulator_setup}.

We ran either one or two participants at the same time. To minimize participant interactions and crosstalk, participants wore Sony WH-1000XM4 headsets with separate pluggable V-Moda BoomPro Microphones and were separated by sound-dampening barriers in the room. (We were not able to use the noise-canceling feature of the headsets due to the need to only have wired connections to the computer). The headset microphones were used to collect verbal responses to auditory-response tasks and to enable communication between the participant and the research assistant. Research assistants ran sessions from a control room and were able to speak with participants if needed and remotely teleport a participant's vehicle if they became stuck.

\subsection{Simulation and Scenario Design}
The CARLA simulated town included residential and arterial roads, buildings, trees, and other scenery, automated traffic behavior, and parked cars (both orthogonal and parallel). 

Our focus was on broad impairment detection -- specifically on the moments when an impaired driver encounters an unexpected event, cannot respond in time, and crashes. Prior work surveying nine impairment domains (alcohol intoxication, cognitive distraction, cognitive load, fatigue, frustration, readiness, stress, sudden sickness, and visual distraction)~\cite{battel2026inconsistent} identified common study methodologies and inconsistencies across them. Building on that survey, we selected the impairments studied here (“impairments”), the methods used to induce them (“inductions”), and the hazards embedded in our scenarios (“hazards”).

\subsection{Study Procedure}
\begin{figure*}[t]
    \centering
    \includegraphics[width=\linewidth]{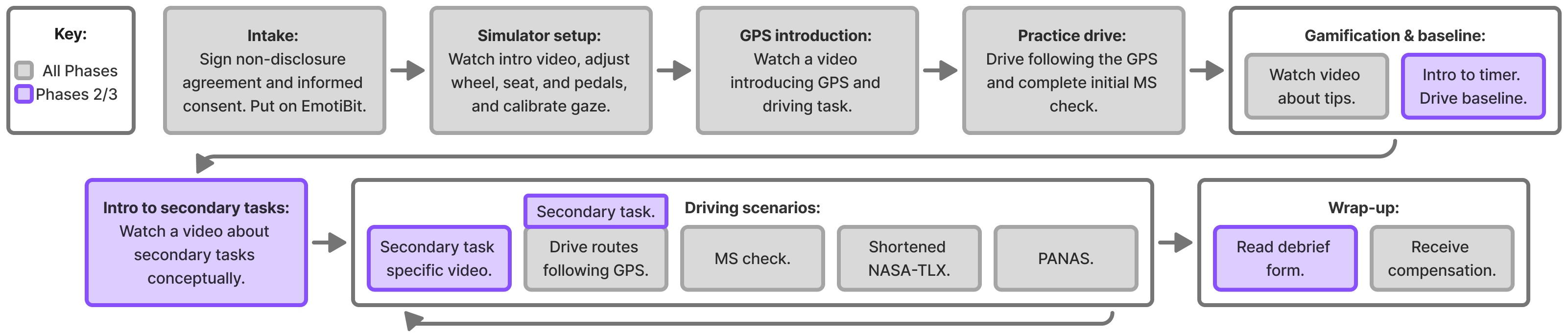}
    \caption{Overview of the study procedure with components from all Phases in gray boxes and components from Phases 2 and 3 (with impairments) in purple boxes.}
    \label{fig:procedure}
\end{figure*}

The following procedure, which can be seen in \Cref{fig:procedure}, was used for the data collection:
\begin{enumerate}
    \item Intake: Participants sign a non-disclosure agreement and informed consent using Qualtrics, then fill out the pre-drive survey on an iPad. An EmotiBit wristband is fitted.
    \item Simulator setup: Participants watch a simulator intro video, adjust the seat, wheel, and pedals.
    \item Gaze calibration: Participants conduct the Tobii on-screen eye-tracker calibration. After this point, participants interact with Research Assistants (RAs) remotely through their headset.
    \item GPS introduction: A video introduces the iPad-based GPS interface and driving task.
    \item Practice drive: Participants watch a driving-task intro video, then complete a practice drive with RA feedback and a motion sickness check.
    \begin{enumerate}
        \item Gamification and Baseline drive: Participants then watch a video introducing the monetary tips and delivery driver scenario. In Phases 2 and 3, the video includes an introduction to the GPS countdown timer and they then complete a 10-minute baseline drive with no inductions.
        \item Introduction to secondary tasks: In Phases 2 and 3, participants watch a video introducing the concept of secondary tasks while driving.
    \end{enumerate}
    \item Driving scenario routes: Participants run the counterbalanced scenario routes. Scenario order comes from a randomization table specific to each phase. Scenarios are interspersed with motion sickness checks, post-drive surveys, and pre-drive videos.
    \begin{enumerate}
        \item Motion sickness checks and mid-drive surveys: After each driving scenario, participants rate motion sickness and complete a mid-drive survey including a shortened NASA-TLX, KSS, and PANAS scale.
        \item Pre-drive videos (Phases 2 and 3 only): Before their next driving scenario, participants watch an instructional video introducing the induction - often a secondary task.
    \end{enumerate}
    \item Compensation and Debrief: RAs re-enter the room, remove the EmotiBit, provide the debrief form (only necessary during Phases 2 and 3 due to the use of audio to induce frustration and stress), and answer questions. Payment was \$170 plus the route bonus for the experimental group and up to \$30 bonus for route tips (see Phase 2 and Phase 3 setups below), plus extra pay for any time past the scheduled session.
\end{enumerate}

\subsubsection{Phase 1: Naturalistic Driving}
This phase consisted of five scenario tables, which each had four driving `loops,' or sets of routes. There were five total loops, so each scenario table started with a different loop and was missing one loop. This phase did include monetary tips but did not include a countdown timer.

\begin{table}[t]
\centering
\small
\setlength{\tabcolsep}{3.5pt}
\begin{tabular}{@{}p{3.2cm}cc@{}}\toprule
\textbf{Task} & \textbf{Drives} & \textbf{Hours} \\
\midrule
\quad music                  & 57 & 4.82 \\
\quad $n$-back               & 49 & 4.83 \\
\quad automated phone call   & 48 & 8.49 \\
\quad scene search           & 44 & 5.86 \\
\quad baby crying            & 44 & 3.65 \\
\quad radio only             & 35 & 3.25 \\
\quad texting                & 35 & 3.06 \\
\quad radio comprehension    & 30 & 5.34 \\
\quad forced reroute$^{*}$   & 20 & 3.76 \\
\quad visual sink$^{*}$      & 19 & 3.36 \\
\bottomrule
\end{tabular}
\\[2pt]
{\footnotesize $^{*}$ Phase 3 only}
\caption{Hours by induction.}
\label{tab:impaired-by-task}
\end{table}

\subsubsection{Phase 2: Impaired Driving}
The addition of the GPS app countdown timer (to place a time limit on the drive, after which tips are deducted) in this Phase was intended to encourage the participant to continue driving while doing their secondary task, as participants may have compensated for their impaired state by slowing down or stopping if no time pressure existed. For a breakdown of hours of driving by induction, see \Cref{tab:impaired-by-task}.

Due to study engineering constraints, Phase 2 was collected in two batches, with differences outlined below. 

\textbf{Batch 1:}
Batch 1 had three scenario tables across 16 completed participants. These scenario tables improved on the ones from Phase 1 and contained a greater variety of driving routes and a more complex, counterbalanced randomization order of five distinct driving scenarios, four of which had the impairments listed below.

\begin{figure*}[t]
    \centering
    \includegraphics[width=\linewidth]{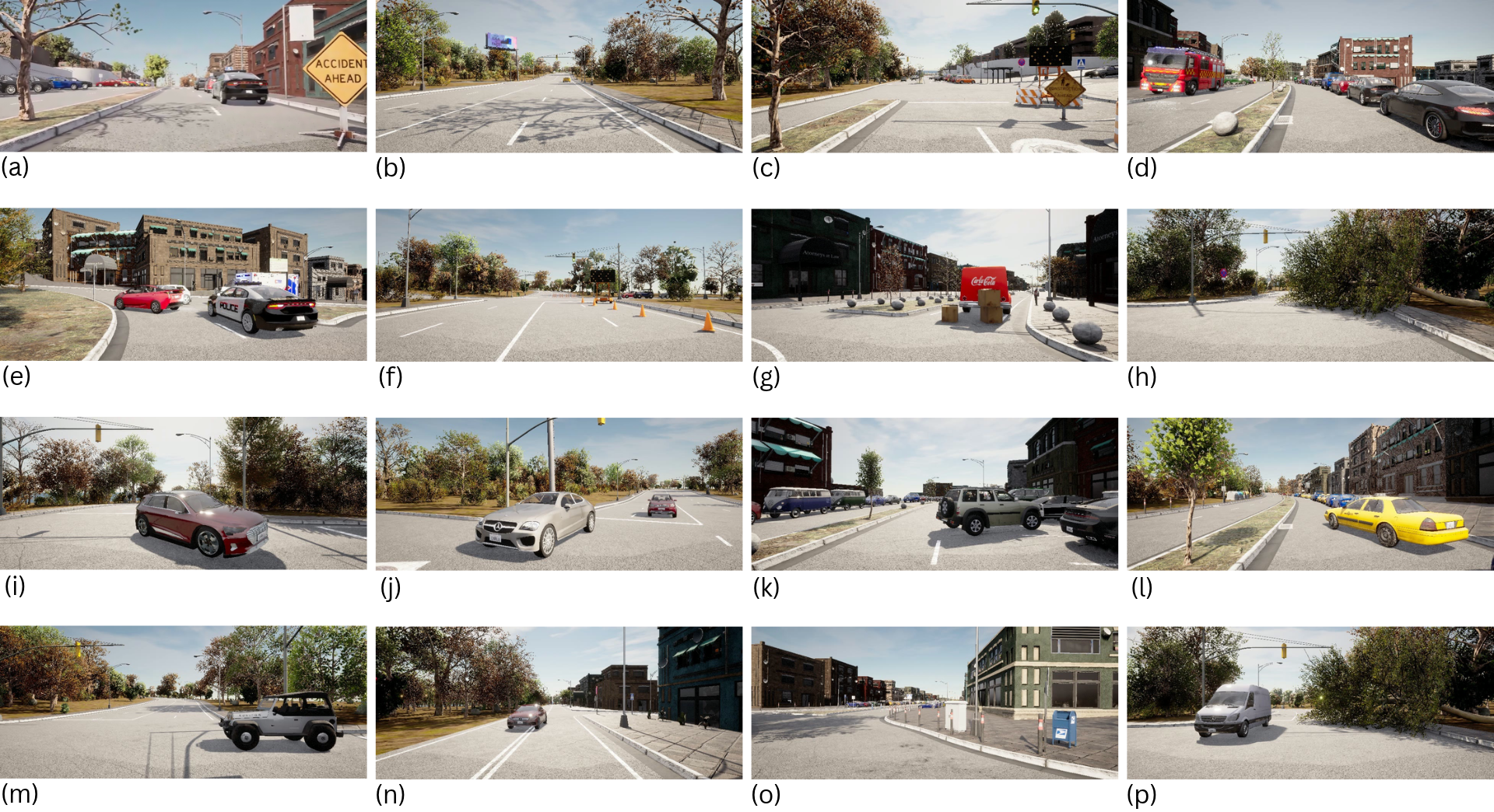}
    \caption{Images of CARLA visual sink props (a-d), forced reroute props (e-h), triggered hazards (i-n), mailbox from scene search (o), and a hazard alongside a forced reroute (p). Specific labels are as follows: (a) accident on the side of the road with police car; (b) billboard; (c) construction sign with cones; (d) emergency vehicles parked on the sidewalk; (e) accident in roundabout with a police car and ambulance; (f) arrow board and set of cones ending the right lane and blocking lane at the far side of the intersection; (g) `Carla Cola' truck with boxes blocking roundabout exit; (h) downed tree blocking the road at the far side of the intersection; (i) cut-off turn hazard; (j) ego vehicle turn hazard; (k) orthogonal parked pop-out car; (l) parallel parked pop-out car; (m) red light runner; (n) reckless driver; (o) blue mailbox from scene search task and (p) cut-off turn hazard triggered near downed tree forced reroute.}
    \label{fig:carla_scenes}
\end{figure*}

Batch 1 contained the first set of participants to experience impairments, with the following inductions:
\begin{enumerate}
    \item Automated phone call: Consisted of an 8-minute and 45-second fake phone call in which a robotic voice would inform the participant of a customer complaint with a previous order and repeatedly put them on hold. This was categorized as continuous and intended to produce cognitive distraction and frustration.
    \item N-back: Consisted of a 1-back task where the participant would listen to a string of numbers and repeat the number before the most recent one. This was categorized as continuous and intended to produce cognitive distraction.
    \item Baby crying: Consisted of a 4-minute clip of a baby crying playing through the participant's headset. This was categorized as continuous and intended to produce cognitive distraction, frustration, and/or stress.
    \item Scene search (mailbox): Consisted of a number of blue mailboxes that were placed around the town in CARLA. Participants were informed to scan the scene while driving and press a button behind the steering wheel whenever one was spotted. In this initial phase, we did not save the participant's button presses. This was categorized as continuous and intended to produce visual distraction, though an increase in cognitive load is to be expected. See \Cref{fig:carla_scenes}(o) for an image of the mailbox in the scene.

\end{enumerate}

\textbf{Batch 2: }
Batch 2 had the same three scenario tables, as well as four more (for a total of seven) across 19 completed participants. The new scenario tables contained a greater variety of driving routes and further changes to the complexity of the randomization with a total of 24 new driving scenarios across the four new scenario tables. This was done and balanced in part to strengthen the coverage of routes across the map. Three participants in Batch 2 also experienced our `red light runner' hazard, in which a vehicle ran a red light from the left or the right in front of the ego vehicle.

In addition to the inductions included for Batch 1, Batch 2 also incorporated the following inductions:
\begin{enumerate}
    \item Music: Consisted of 2 songs of the participant's choice being played through their headphones. This was categorized as passive but could be expected to induce minor cognitive distraction.
    \item Radio segment: Consisted of a 5-minute, 19-second audio clip about the history of electric vehicles. This was categorized as passive but could be expected to induce minor cognitive distraction.
    \item Radio segment + comprehension: Consisted of a 5-minute, 37-second audio clip about the English Channel, followed by a 4 minute recording of multiple choice questions about the clip. This was categorized as continuous and intended to induce greater cognitive distraction than the radio segment alone.
    \item Texting: Consisted of 3 texting cases with 3 possible messages that would pop-up with a `ding' on the participant's iPad. In one case, the participant needed to swipe the text to dismiss it, in another they could click one of two responses to dismiss it, and in the final one it would disappear after a few seconds. Messages included the following options: (1) ``Are you going to be on time?", (2) ``I'm waiting at the [landmark]" where landmark was populated with a prop near the delivery point, and (3) ``You've received another 5 star rating!" Messages were not necessarily linked to texting scenario case. Participants were informed to acknowledge aloud having received the text. This was categorized as instantaneous and intended to induce visual distraction and cognitive distraction.
\end{enumerate}
We also added a larger number and variety of locations of blue mailboxes from the scene search task to vary the gaze patterns and allow for a greater number of driving routes to pass by mailboxes, as we were expanding the area of the map coverable by different impairments.

\subsubsection{Phase 3: Impaired Driving with Hazards}
Our goal in Phase 3 was to collect data of specific moments where we knew drivers could be impaired by one of several inductions and at risk of crashing into an unexpected hazard. This Phase contained 22 scenario tables with 142 scenarios and 87 unique delivery routes, across 49 included participants. Scenario tables were balanced for breadth across potential routes on the map, exposure to hazards, and exposure to impairments. 

In Phase 3, we added the following impairment inductions: 
\begin{enumerate}
    \item `Visual sink': See \Cref{fig:carla_scenes} for images and a caption with more detail. Consisted of 8 locations across the map that had unique or interesting props, which were categorized as instantaneous and intended as a visual distraction and may have increased cognitive load due to complexity of the scene. These included the following: (1) billboard, (2) emergency vehicles on a sidewalk, (3) accident on the side of the road, (4) construction sign with cones, (5) accident in a roundabout, (6) `carla cola' truck with boxes, (7) construction with arrow board and set of cones, and (8) downed tree. 
    \item `Forced reroutes': Visual sinks 5-8 were co-categorized as forced reroutes, which would in some routes block the participant from following the GPS route, and in others serve as alternative visual sink. Forced reroutes were categorized as instantaneous and intended to induce cognitive load due to the complexity of the scene, and potentially frustration and stress due to an impact on drive time.
\end{enumerate}

\begin{figure*}[ht]
    \centering
    \begin{subfigure}[t]{0.65\textwidth}
        \centering
        \vspace{0pt}%
        \includegraphics[width=\linewidth]{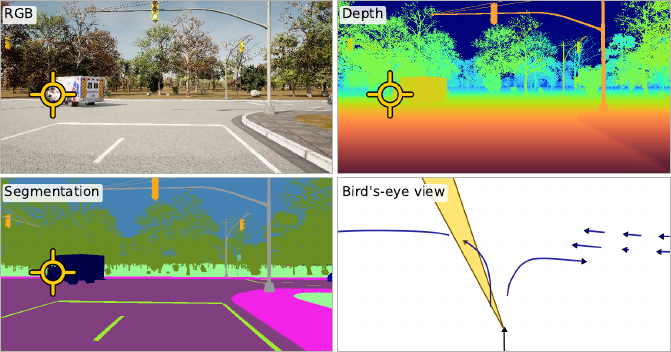}
        \caption{The three released video modalities (color, depth, and segmentation) and trajectories. The gaze is overlaid on each as yellow crosshairs and a view triangle.}
        \label{fig:data_records_frame}
    \end{subfigure}\hfill
    \begin{subfigure}[t]{0.33\textwidth}
        \centering
        \vspace{0pt}%
        \includegraphics[width=\linewidth]{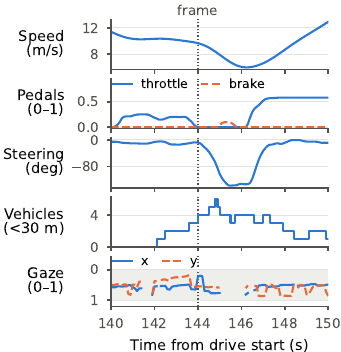}
        \caption{Temporal plots of control and gaze modalities in the dataset. Dotted line shows current time for \subref{fig:data_records_frame}.}
        \label{fig:data_records_signals}
    \end{subfigure}
    \caption{All released modalities shown for a ten second clip. Not depicted: impairment labels, surveys, pupil diameter.}
    \label{fig:data_records}
\end{figure*}

\begin{table*}[ht]
    \centering
    \small
    \begin{tabular}{l l r p{0.46\textwidth}}
        \toprule
        Modality & Source & Hz & Contents (units) \\
        \midrule
        Scene video, RGB & CARLA forward camera & 10 & $960 \times 400$ frames (MJPEG) \\
        Scene video, depth & CARLA depth camera & 10 & $960 \times 400$ frames (lossless); depth (max of 1000\,m) per pixel \\
        Scene video, segmentation & CARLA semantic camera & 10 & $960 \times 400$ frames (lossless); CARLA semantic class per pixel \\
        \midrule
        Eye gaze & Tobii Pro Spark & 60 & per eye: gaze point on display (normalized), gaze origin (mm), pupil diameter (mm), validity \\
        Vehicle states & CARLA vehicle telemetry & 10 & pose (m, quaternion), bounding box, actor type \\
        Driver controls & Fanatec wheel and pedals & 10 & steering-wheel angle (deg), throttle and brake pedal (0--1) \\
        \midrule
        Impairment labels & protocol per drive & -- & induction task and impairment (baseline, passive, continuous, instantaneous) \\
        Surveys & after each drive & -- & shortened NASA-TLX, KSS, PANAS, motion-sickness check \\
        \bottomrule
    \end{tabular}
    \caption{Modalities released with \dsname{}. Signals are time-aligned based on a scenario start message.}
    \label{tab:data_records}
\end{table*}

Phase 3 included 21 hazards; every scenario table started with a 10-minute baseline (no impairment) and included five impaired driving scenarios and a separate 5-minute baseline. Each scenario was five to ten minutes long, contained up to three delivery routes, contained up to 2 hazards in relevant locations, contained up to 2 visual sinks, and contained up to one forced reroute. The order of impairments and the 5-minute baseline were randomized.
The following is a list of the programmed hazards we included (see \Cref{fig:carla_scenes} for images):
\begin{enumerate}
    \item Four red light runners -- defined as a vehicle running a red light from the ego driver's right or left in front of their vehicle.
    \item Five orthogonal parked pop-out cars -- defined as an orthogonally-parked car pulling out suddenly in front of the ego vehicle. The programmed pop-out behavior included a short reversal early, a brake for a few seconds, and then completion of the pop-out. This was intended to allow the ego driver to notice that the car might pull out if they were paying attention.
    \item  Five parallel parked pop-out cars -- defined as a parallel-parked car pulling out suddenly in front of the ego vehicle. Similar to orthogonal pop-outs, the parallel pop-out would move back and forth and poke its front out with enough time for the driver to notice and react if they were paying attention.
    \item One reckless driver -- defined as a vehicle swerving and oncoming in the wrong lane before entering the proper lane.
    \item Three cut-off turn hazards -- defined as a separate vehicle turning suddenly at a light in front of the ego vehicle.
    \item  Three ego vehicle turn hazards -- defined as an oncoming car at a green light where the ego driver needed to turn left, which caused the potential for a collision if the ego vehicle turned into the oncoming car.
\end{enumerate}

In some cases, hazards were deliberately triggered at locations associated with visual sinks or forced reroutes, so that impairment (i.e., cognitive load from a more difficult scene, or visual distraction) would have more opportunity to produce a collision. See \Cref{fig:carla_scenes}(p) for an example of such a scenario.

\begin{table}[t]
\centering
\small
\setlength{\tabcolsep}{3.5pt}
\begin{tabular}{@{}p{3.2cm}cccc@{}}\toprule
 & \textbf{P1}& \textbf{P2}& \textbf{P3}& \textbf{Total} \\
\midrule
\multicolumn{5}{@{}l}{\textit{Dataset}} \\
\midrule
\quad Enrolled Subjects           & 33 & 41 & 54 & 128 \\
\quad Completed Subjects          & 25 & 35 & 49 & 109 \\
\quad Hours practice       & 2.5           & 3.1           & 4.0           & 9.6 \\
\quad Hours baseline       & 28.6          & 6.8           & 11.9          & 47.3 \\
\quad Hours continuous     & 0             & 16.0          & 12.2          & 28.2 \\
\quad Hours passive        & 0             & 1.9           & 6.1           & 8.0 \\
\quad Hours instantaneous  & 0             & 1.0           & 9.2           & 10.2 \\
\quad \textbf{Total hours} & \textbf{31.1} & \textbf{28.8} & \textbf{43.4} & \textbf{103.3} \\

\midrule
\multicolumn{5}{@{}l}{\textit{Demographics}} \\
\midrule
\quad Age, range          & 22-66& 22-64& 21-63& 21-66\\
 \quad Age, mean& 39.6& 36.4& 37.7&37.7\\
\quad Age, SD& 12.4& 11.6& 9.0& 10.7\\
\midrule
\quad Female, $n$ (\%)    & 16 (64) & 20 (57.1)& 27 (55.1)& 63 (57.8)\\
\quad Male, $n$ (\%)      & 9 (36)& 14 (40)& 22 (44.9)& 45 (41.3)\\
\quad Non-binary, $n$ (\%)& 0 (--)& 1 (2.9)& 0 (--)& 1 (0.9)\\
\midrule
 \quad White, $n$ (\%)& 17 (68) & 27 (77.1) & 38 (77.6) & 82 (75.2)\\
 \quad Hispanic or\\ \quad \quad Latino, $n$ (\%)& 4 (16) & 1 (2.9) & 5 (10.2) & 10 (9.2)\\
 \quad Asian, $n$ (\%)& 3 (12) & 3 (8.6) & 3 (6.1) & 9 (8.3)\\
 \quad Black or African \\ \quad \quad American, $n$ (\%)& 1 (4) & 4 (11.4) & 3 (6.1) & 8 (7.3)\\
\end{tabular}
\caption{Participant demographics and dataset statistics. P1-P3 refer to Phases.}
\label{tab:demographics}
\end{table}

\begin{table*}[p]  %
  \centering
  \renewcommand{\arraystretch}{0.9}
  \footnotesize
  \setlength{\tabcolsep}{4pt}
  \resizebox{\ifdim\width>\textwidth\textwidth\else\width\fi}{!}{%
  \begin{tabular}{@{}lr crr crr@{}}
    \toprule
    \multicolumn{8}{@{}l}{\textit{(a) Workload and sleepiness}} \\[2pt]
    & & \multicolumn{3}{c}{NASA-TLX} & \multicolumn{3}{c}{KSS} \\
    \cmidrule(lr){3-5} \cmidrule(l){6-8}
    Task & $n$ & $M_\text{diff}$ [95\% CI] & $p$ & $d_z$ & $M_\text{diff}$ [95\% CI] & $p$ & $d_z$ \\
    \midrule
    N-back & 48 & \textbf{4.97 [4.23, 5.72]} & \textbf{$<$.001} & \textbf{1.94} & $-$0.11 [$-$0.30, 0.07] & $>$.99 & $-$0.18 \\
    Radio comprehension & 31 & \textbf{2.51 [1.56, 3.45]} & \textbf{$<$.001} & \textbf{0.98} & 0.13 [$-$0.07, 0.32] & $>$.99 & 0.24 \\
    Forced reroute & 22 & 1.02 [0.24, 1.80] & .106 & 0.58 & 0.07 [$-$0.13, 0.27] & $>$.99 & 0.15 \\
    Phone call & 48 & 0.62 [$-$0.24, 1.48] & .973 & 0.21 & 0.16 [$-$0.05, 0.36] & $>$.99 & 0.22 \\
    Scene search & 44 & 0.57 [$-$0.19, 1.34] & .973 & 0.23 & 0.01 [$-$0.20, 0.22] & $>$.99 & 0.02 \\
    Radio only & 35 & 0.55 [$-$0.19, 1.29] & .973 & 0.25 & 0.13 [$-$0.10, 0.36] & $>$.99 & 0.19 \\
    Baby crying & 45 & 0.55 [$-$0.38, 1.47] & .973 & 0.18 & $-$0.04 [$-$0.26, 0.17] & $>$.99 & $-$0.06 \\
    Texting & 35 & 0.30 [$-$0.65, 1.26] & $>$.99 & 0.11 & 0.10 [$-$0.11, 0.31] & $>$.99 & 0.16 \\
    Visual sink & 20 & $-$0.09 [$-$1.24, 1.06] & $>$.99 & $-$0.04 & $-$0.05 [$-$0.25, 0.15] & $>$.99 & $-$0.12 \\
    Music control & 56 & $-$0.40 [$-$1.01, 0.22] & .973 & $-$0.17 & $-$0.03 [$-$0.20, 0.15] & $>$.99 & $-$0.04 \\
    \midrule
    \multicolumn{8}{@{}l}{\textit{(b) Affect}} \\[2pt]
    & & \multicolumn{3}{c}{PANAS negative} & \multicolumn{3}{c}{PANAS positive} \\
    \cmidrule(lr){3-5} \cmidrule(l){6-8}
    Task & $n$ & $M_\text{diff}$ [95\% CI] & $p$ & $d_z$ & $M_\text{diff}$ [95\% CI] & $p$ & $d_z$ \\
    \midrule
    N-back & 48 & \textbf{1.11 [0.56, 1.67]} & \textbf{.002} & \textbf{0.59} & $-$0.90 [$-$1.98, 0.17]$^\dagger$ & .879 & $-$0.25 \\
    Radio comprehension & 31 & \textbf{0.73 [0.28, 1.17]} & \textbf{.021} & \textbf{0.60} & 0.05 [$-$0.88, 0.98] & $>$.99 & 0.02 \\
    Forced reroute & 22 & 0.55 [0.02, 1.07] & .308 & 0.46 & 0.23 [$-$0.32, 0.78] & $>$.99 & 0.18 \\
    Phone call & 48 & 0.82 [$-$0.14, 1.78] & .544 & 0.25 & $-$0.61 [$-$1.36, 0.14] & .879 & $-$0.24 \\
    Scene search & 44 & $-$0.07 [$-$0.54, 0.40] & $>$.99 & $-$0.04 & $-$0.13 [$-$1.26, 1.00]$^\dagger$ & $>$.99 & $-$0.03 \\
    Radio only & 35 & 0.03 [$-$0.25, 0.30] & $>$.99 & 0.04 & 0.09 [$-$0.69, 0.86] & $>$.99 & 0.04 \\
    Baby crying & 45 & $-$0.11 [$-$0.59, 0.36] & $>$.99 & $-$0.07 & $-$0.28 [$-$1.38, 0.83] & $>$.99 & $-$0.08 \\
    Texting & 35 & $-$0.09 [$-$0.42, 0.25] & $>$.99 & $-$0.09 & 0.51 [$-$0.29, 1.32] & $>$.99 & 0.22 \\
    Visual sink & 20 & 0.07 [$-$0.42, 0.57] & $>$.99 & 0.07 & 0.28 [$-$0.33, 0.88] & $>$.99 & 0.21 \\
    Music & 56 & \textbf{$-$0.34 [$-$0.55, $-$0.13]} & \textbf{.021} & \textbf{$-$0.42} & 0.61 [0.14, 1.08]$^\dagger$ & .122 & 0.35 \\
    \bottomrule
  \end{tabular}}
  \caption{Overall measures. Paired $t$-tests of each task versus each participant's mean baseline score ($N = 84$ participants). $M_\text{diff}$ = task minus baseline; $n$ = participants with both scores ($\text{df} = n - 1$). $p$-values are Holm-adjusted across the ten tasks within each measure; bold = $p < .05$. $d_z$ = mean difference / SD of differences. NASA-TLX: 0--20; KSS: 1--9; PANAS: 5--25. $^\dagger$One fewer participant than the $n$ shown (a skipped item).}
  \label{tab:survey_overall}
  \vspace{1em}
  \centering
  \footnotesize
   \setlength{\tabcolsep}{4pt}
  \resizebox{\ifdim\width>\textwidth\textwidth\else\width\fi}{!}{%
  \begin{tabular}{@{}lr crr crr crr@{}}
    \toprule
    \multicolumn{11}{@{}l}{\textit{(a) Demand subscales}} \\[2pt]
    & & \multicolumn{3}{c}{Mental demand} & \multicolumn{3}{c}{Physical demand} & \multicolumn{3}{c}{Temporal demand} \\
    \cmidrule(lr){3-5} \cmidrule(lr){6-8} \cmidrule(l){9-11}
    Task & $n$ & $M_\text{diff}$ [95\% CI] & $p$ & $d_z$ & $M_\text{diff}$ [95\% CI] & $p$ & $d_z$ & $M_\text{diff}$ [95\% CI] & $p$ & $d_z$ \\
    \midrule
    N-back & 48 & \textbf{6.92 [5.59, 8.24]} & \textbf{$<$.001} & \textbf{1.51} & \textbf{2.73 [1.75, 3.71]} & \textbf{$<$.001} & \textbf{0.81} & \textbf{4.68 [3.36, 6.00]} & \textbf{$<$.001} & \textbf{1.03} \\
    Radio comprehension & 31 & \textbf{3.76 [2.21, 5.30]} & \textbf{$<$.001} & \textbf{0.89} & 0.71 [$-$0.25, 1.67] & .573 & 0.27 & 1.37 [0.17, 2.57] & .209 & 0.42 \\
    Forced reroute & 22 & 1.16 [$-$0.10, 2.42] & .556 & 0.41 & 0.86 [$-$0.06, 1.79] & .326 & 0.41 & \textbf{2.34 [1.01, 3.67]} & \textbf{.013} & \textbf{0.78} \\
    Phone call & 48 & 0.93 [$-$0.27, 2.12] & .700 & 0.23 & \textbf{1.26 [0.42, 2.10]} & \textbf{.034} & \textbf{0.43} & $-$0.26 [$-$1.57, 1.05] & $>$.99 & $-$0.06 \\
    Scene search & 44 & 1.17 [$-$0.26, 2.60] & .700 & 0.25 & \textbf{1.44 [0.57, 2.32]} & \textbf{.016} & \textbf{0.50} & $-$0.53 [$-$1.72, 0.65] & $>$.99 & $-$0.14 \\
    Radio only & 35 & 0.86 [$-$0.32, 2.03] & .700 & 0.25 & 0.83 [0.21, 1.45] & .073 & 0.46 & $-$0.26 [$-$1.52, 1.00] & $>$.99 & $-$0.07 \\
    Baby crying & 45 & 0.59 [$-$0.79, 1.97] & $>$.99 & 0.13 & 0.83 [0.08, 1.58] & .180 & 0.33 & 0.19 [$-$1.14, 1.52] & $>$.99 & 0.04 \\
    Texting & 35 & 1.10 [$-$0.22, 2.42] & .700 & 0.29 & 0.53 [$-$0.42, 1.48] & .799 & 0.19 & $-$0.73 [$-$2.02, 0.56] & $>$.99 & $-$0.19 \\
    Visual sink & 20 & $-$0.45 [$-$1.97, 1.07] & $>$.99 & $-$0.14 & 0.45 [$-$0.59, 1.49] & .799 & 0.20 & 0.12 [$-$1.08, 1.33] & $>$.99 & 0.05 \\
    Music & 56 & $-$0.40 [$-$1.33, 0.52] & $>$.99 & $-$0.12 & 0.06 [$-$0.49, 0.62] & .822 & 0.03 & $-$0.08 [$-$1.01, 0.85] & $>$.99 & $-$0.02 \\
    \midrule
    \multicolumn{11}{@{}l}{\textit{(b) Performance, effort and frustration}} \\[2pt]
    & & \multicolumn{3}{c}{Performance (reversed)} & \multicolumn{3}{c}{Effort} & \multicolumn{3}{c}{Frustration} \\
    \cmidrule(lr){3-5} \cmidrule(lr){6-8} \cmidrule(l){9-11}
    Task & $n$ & $M_\text{diff}$ [95\% CI] & $p$ & $d_z$ & $M_\text{diff}$ [95\% CI] & $p$ & $d_z$ & $M_\text{diff}$ [95\% CI] & $p$ & $d_z$ \\
    \midrule
    N-back & 48 & \textbf{4.97 [3.85, 6.08]} & \textbf{$<$.001} & \textbf{1.29} & \textbf{5.95 [4.64, 7.26]} & \textbf{$<$.001} & \textbf{1.32} & \textbf{4.60 [3.18, 6.03]} & \textbf{$<$.001} & \textbf{0.94} \\
    Radio comprehension & 31 & \textbf{4.13 [2.43, 5.83]} & \textbf{$<$.001} & \textbf{0.89} & \textbf{2.95 [1.92, 3.98]} & \textbf{$<$.001} & \textbf{1.05} & \textbf{2.13 [0.78, 3.48]} & \textbf{.022} & \textbf{0.58} \\
    Forced reroute & 22 & 1.39 [0.11, 2.67] & .281 & 0.48 & 0.25 [$-$1.02, 1.52] & $>$.99 & 0.09 & 0.11 [$-$1.13, 1.36] & $>$.99 & 0.04 \\
    Phone call & 48 & 0.97 [$-$0.20, 2.13] & .706 & 0.24 & $-$0.19 [$-$1.58, 1.20] & $>$.99 & $-$0.04 & 1.02 [$-$0.47, 2.51] & $>$.99 & 0.20 \\
    Scene search & 44 & 1.03 [$-$0.22, 2.29] & .706 & 0.25 & 1.03 [$-$0.34, 2.41] & .816 & 0.23 & $-$0.70 [$-$1.79, 0.38] & $>$.99 & $-$0.20 \\
    Radio only & 35 & 0.90 [$-$0.33, 2.13] & .730 & 0.25 & 0.47 [$-$0.72, 1.66] & $>$.99 & 0.14 & 0.49 [$-$0.71, 1.68] & $>$.99 & 0.14 \\
    Baby crying & 45 & $-$0.26 [$-$1.63, 1.12] & $>$.99 & $-$0.06 & $-$0.61 [$-$1.97, 0.75] & $>$.99 & $-$0.14 & \textbf{2.53 [1.04, 4.03]} & \textbf{.013} & \textbf{0.51} \\
    Texting & 35 & 0.99 [$-$0.68, 2.65] & .953 & 0.20 & $-$0.09 [$-$1.40, 1.23] & $>$.99 & $-$0.02 & 0.01 [$-$1.14, 1.17] & $>$.99 & 0.00 \\
    Visual sink & 20 & 0.42 [$-$1.87, 2.72] & $>$.99 & 0.09 & $-$1.20 [$-$2.59, 0.19] & .613 & $-$0.40 & 0.10 [$-$1.84, 2.04] & $>$.99 & 0.02 \\
    Music control & 56 & 0.26 [$-$0.97, 1.49] & $>$.99 & 0.06 & $-$0.97 [$-$1.83, $-$0.12] & .216 & $-$0.30 & \textbf{$-$1.24 [$-$2.03, $-$0.46]} & \textbf{.020} & \textbf{$-$0.42} \\
    \bottomrule
  \end{tabular}}
  \caption{NASA-TLX subscales (0--20). Paired $t$-tests of each task versus each participant's mean baseline score ($N = 84$ participants). $M_\text{diff}$ = task minus baseline; $n$ = participants with both scores ($\text{df} = n - 1$). $p$-values are Holm-adjusted across the ten tasks within each measure; bold = $p < .05$. $d_z$ = mean difference / SD of differences. Performance is reverse-scored, so higher = worse perceived performance (more workload) on every subscale.}
  \label{tab:tlx_subscales}
\end{table*}

\subsection{Participant Statistics and Demographics}
Table \ref{tab:demographics} provides an overview of participant statistics and demographics. Age, gender, and race statistics in the table are counted from the completed participant pool and do not include dropouts.

Participant dropouts were typically due to motion sickness ($n=16$ across all phases), but two participants were discounted completely due to driving behavior (many missed turns, difficulty following GPS, erratic driving behavior), and one due to technical issues with the simulator. Of the remaining participants, some participant data was incomplete and the subsequent dataset is missing such parts. Scenarios leading to this outcome included: driving too slowly and not finishing all scenarios in the session, technical issues leading to restarted or missed sections of loops in the scenario, reactive driving and crashing into stationary objects leading to sections of driving not being included, non-catastrophic technical issues (such as audio files not playing) during parts of scenarios, and catastrophic technical issues that occurred later in a participant's session and ended the session early.

\subsection{Data Records}
\label{sec:data_records}
\dsname{} is organized by participant and drive. Each drive can contain multiple GPS delivery routes, with a single type of impairment condition induced. The raw data has been preprocessed into time-synchronized modalities outlined in \Cref{fig:data_records} and \Cref{tab:data_records}, including video, trajectory, and control data at 10 Hz, as well as gaze data at 60 Hz. The entire-drive impairment labels and post-drive surveys are also provided. Modalities collected but not released to preserve participant anonymity include driver-facing video, EmotiBit physiological data, and audio recordings. Further details about directory structure will be added upon public dataset release.

\subsection{\texorpdfstring{Further Survey Analyses}{Further Survey Analyses}}
\label{sec:tech_validation}
To check that the inductions were noticeable, we compared all post-drive surveys (PANAS, TLX, and KSS) for each task against each participant's mean baseline score with paired $t$-tests and Holm correction. Full results can be found in \Cref{tab:survey_overall} and \Cref{tab:tlx_subscales}. TLX subscales findings are discussed in \Cref{sec:dataset}. For the KSS measure, there were no significant differences between any of the inductions and baselines. On the PANAS, the N-back and radio comprehension tasks elicited more negative affect, whereas the music control condition resulted in reduced negative affect. No significant differences were found on the PANAS positive affect scale.

\section{\texorpdfstring{Evaluation Protocol}{Evaluation Protocol}}
\label{sec:eval_protocol}

\subsection{\texorpdfstring{Metrics}{Metrics}}
\label{sec:metrics}
\paragraph{Clip scores.}
Impairment predictions for each clip are generated by masking all impairment tokens and providing all other modalities as clean input. Each of the $k$ impairment modality binary heads produces an output at 10 Hz. We average all of these output logits over the clip to produce a final set of $s_k$ logits per clip. Because each head has separate losses, we Z-normalize the outputs by subtracting the mean and dividing by the standard deviation of the held-out validation baseline clips. This provides a normalized set of $z_k$ logits. A clips overall impairment score is the maximum over all the heads: $z_{\text{any}} = \max_k z_k$. No test labels are used to calibrate the heads.

\paragraph{P-AUC.}
For each continuous impairment $k$, the P-AUC compares the $z_{\text{any}}$ between clips of impairment $k$ and baseline clips. In \dsname only the location-matched pairs are compared (using the Mann--Whitney estimate of AUC), while the AUC for IDD is calculated using the standard unmatched method. The final P-AUC value is the mean of the five per-impairment AUCs, so each impairment counts equally regardless of how many pairs it has.

\paragraph{I-AUC.}
The I-AUC is designed to show the overall differentiation capability of the model between continuous impairment types. For each of the ten possible combinations of impairment $x$ versus impairment $y$, we contrast the clips of each type and calculate the difference between the heads as $d_{xy} = z_y - z_x$. In \dsname only the location-matched pairs are compared (using the Mann--Whitney estimate of AUC), while the AUC for IDD is calculated using the standard unmatched method. The final I-AUC value is the mean of the ten impairment-versus-impairment AUCs, so each combination counts equally regardless of how many pairs it has.

\begin{table}[t]
    \centering
    \small
    \setlength{\tabcolsep}{2pt}
    \fittable{\begin{tabular}{l r r r r r r r}
        \toprule
         & Scene & N-back & Radio & Phone & Baby & Base. & Clips \\
        \midrule
        Scene & -- &  &  &  &  &  & 3{,}392 \\
        N-back & 622 & -- &  &  &  &  & 2{,}326 \\
        Radio & 1{,}472 & 861 & -- &  &  &  & 2{,}860 \\
        Phone & 2{,}913 & 1{,}392 & 2{,}148 & -- &  &  & 4{,}807 \\
        Baby & 523 & 314 & 758 & 1{,}246 & -- &  & 1{,}855 \\
        Baseline & 5{,}666 & 3{,}971 & 5{,}479 & 10{,}020 & 5{,}793 & -- & 13{,}905 \\
        \bottomrule
    \end{tabular}}
    \caption{Location-matched pairs for each impairment-versus-baseline and impairment-versus-impairment combination (lower triangle only, since the matrix is symmetric). The last column is the number of clips per class. Clips can be reused in multiple different comparison pairs.}
    \label{tab:paired_imbalance}
\end{table}

\subsection{\texorpdfstring{\dsname Location-Matched Pairs}{\dsname Location-Matched Pairs}}
\label{sec:matched_pairs}
\paragraph{Anchored clips.}
The \dsname validation and test clips are chosen to begin at fixed locations on that map rather than at fixed times into the drive. We quantize all ego pose data using a 100\,m grid and further subdivide into four $90^\circ$ heading bins. For each drive, we start a 30\,s clip when the ego enters a new bin. We also require that clips may not be within the first or final 30\,s to avoid potential edge effects. This gives a total of 29{,}145 clips from 441 drives at 425 anchors.

\paragraph{Pair rule.}
While constraining two compared clips to start from the same location reduces the possibility of location leak, because clips continue for 30\,s they may deviate sufficiently in path and still allow for the path to predict impairment. To mitigate this, we implement a pair rule, which requires that clips follow a similar path in order to be considered a valid comparison. To calculate this overlap, we first divide the ego pose data into a more granular 50\,m grid and further subdivide into eight $45^\circ$ heading bins. We take the set of these bins visited by each clip in 30\,s and calculate the coverage as $|A \cap B| / \max(|A|, |B|)$. We retain pairs that have a coverage of at least 0.8. We also restrict pairs to be drawn from the same Phase. The number of matched pairs per class comparison is shown in \Cref{tab:paired_imbalance}. In total, the \dsname location-matched pairs comprise 30,929 impairment-versus-baseline pairs (P-AUC) and 12,249 impairment-versus-impairment pairs (I-AUC). The pairs are unevenly distributed. Phone call has the most clips and therefore the most pairs, while N-back versus baby crying has only 314 pairs making its AUC the least certain.

\begin{table}[t]
    \centering
    \small
    \begin{tabular}{l r r r r}
        \toprule
        & \multicolumn{2}{c}{Pairs} & & \\
        \cmidrule(lr){2-3}
        Pair rule &  B vs. I & I vs. I & P-AUC & I-AUC \\
        \midrule
        None (unpaired) & All & All & 0.602 & 0.810 \\
        Same anchor & 74{,}799 & 32{,}213 & 0.558 & 0.683 \\
        \hspace{1em}+ coverage $\geq 0.8$ & 30{,}929 & 12{,}249 & 0.529 & 0.527 \\
        \bottomrule
    \end{tabular}
    \caption{The location-only baseline under different pair rules. Further restricting pairs reduces the potential location leak in performance, but reduces the pairs available for metric calculation.}
    \label{tab:location_rules}
\end{table}

\paragraph{Effect on location leakage.}
To determine the impact of location leakage, we build a model that only takes coarse location data as an input and predicts impairment. The model takes the 50\,m $\times$ eight-heading cells from the pair rule as input, using a one-hot representation. We use logistic regression and train similarly to \modelname (three training folds, calibrate on validation, score on test fold). \Cref{tab:location_rules} show the performance of the location-only model with different pair rules, showing the tradeoff between reduced potential location leak and the number of pairs for metric calculation.

\section{Further Ablations}

\subsection{Full Modality-Subset Results}
\label{sec:modality_full}
\begin{table}[t]
    \centering
    \small
        \setlength{\tabcolsep}{3pt}\fittable{\begin{tabular}{c c c r r}
        \toprule
        Gaze & Traj. & Video & P-AUC & I-AUC \\
        \midrule
         \checkmark &\checkmark & \checkmark & \textbf{0.706} & 0.728  \\
         \checkmark &        \checkmark &  & 0.688 & 0.718  \\
         \checkmark &         & \checkmark & 0.691 & \textbf{0.731}  \\
         \checkmark &         &  & 0.665 & 0.720  \\
          &        \checkmark & \checkmark & 0.551 & 0.544  \\
          &        \checkmark &  & 0.554 & 0.546  \\
          &         & \checkmark & 0.517 & 0.523  \\
        \bottomrule
    \end{tabular}}
    \caption{Training on a subset of modalities and then testing on impairment classification. A subset of this table is available in the main text in Table~\ref{tab:ablation_main}}
    \label{tab:modality_full}
\end{table}

Table~\ref{tab:modality_full} gives the full results for all seven modality subsets. The main text ablation experiments (Table~\ref{tab:ablation_main}) already shows a comparison between all modalities, gaze only, and trajectory + video. As discussed in the main text, the gaze modality contributes most to the performance. Both trajectory and video boost the P-AUC when added to gaze, but perform near chance on their own.

\subsection{Unsupervised Data}
\label{supp:ablations_unsup}
\begin{table}[t]
    \centering
    \small
        \setlength{\tabcolsep}{3pt}\fittable{\begin{tabular}{l r r}
        \toprule
        Configuration & P-AUC & I-AUC \\
        \midrule
        Drop instantaneous impairments from train & 0.686 & 0.714  \\
        \modelname & 0.706 & 0.728  \\
        Include Phase-1 in train (unlabelled) & \textbf{0.711} & \textbf{0.733}  \\
        \bottomrule
    \end{tabular}}
    \caption{The effects of unlabelled data during training on impairment classification.}
    \label{tab:ablation_unsup}
\end{table}

We wanted to examine the benefit of unsupervised data during training to impairment classification, as the loss also includes reconstruction of trajectory and gaze modalities. \modelname already incorporates unlabelled data from instantaneous impairments clips -- so we first examine removing this data and only using baseline clips and continuous impairments. We next add further data to \modelname -- training with all the Phase~1 data (non-impaired, but treated as unlabelled). With each increase in data, we see a slight improvement in performance at an increased train time. For this reason, we compromised and selected the middle option as \modelname.

\subsection{Masking}
\label{supp:ablations}
\begin{table}[t]
    \centering
    \small
        \setlength{\tabcolsep}{3pt}\fittable{\begin{tabular}{l r r}
        \toprule
        Configuration & P-AUC & I-AUC \\
        \midrule
        \modelname & \textbf{0.706} & 0.728  \\
        Future (trailing) masks only & 0.689 & 0.721  \\
        Impairment always masked & 0.684 & \textbf{0.736}  \\
        No modality dropout & 0.692 & 0.728  \\
        \bottomrule
    \end{tabular}}
    \caption{Masking ablations. We examine the impact of different masking strategies and notice only minor changes in performance.}
    \label{tab:ablation_masking}
\end{table}

Table~\ref{tab:ablation_masking} gives the masking ablations that are not in the main text. We examine the benefit of middle window interpolation in \modelname by instead just using future (trailing) masks for prediction only. We also look to reduce the number of masking configurations considered -- only considering mask combination where impairment is masked and never entirely dropping out modalities. Every masking ablation has a lower P-AUC than \modelname, indicating a slight benefit of the additional masking strategies to generalized impairment detection. However, the difference in performance is mixed and minimal to I-AUC. Overall, it seems most important to have any masking and reconstruction strategy and the model is somewhat robust to different masking techniques.

\section{Qualitative Results}
\label{sec:qualitative}
We release a video along with this supplementary to demonstrate a set of qualitative results. It is available at \url{https://www.youtube.com/watch?v=IvG7YobBwdY}. The sections of the video are as follows:

\begin{itemize}
    \item \emph{Dataset variety.} 12 front-camera clips play at once, showing different event types.
    \item \emph{Released modalities.} The drive of \Cref{fig:data_records} -- showing the released modalities in motion.
    \item \emph{Impairment prediction over time.} A location-matched pair from the evaluation showing the ability of the model to distinguish scene search from a baseline clip with increasing temporal input size.
    \item \emph{Trajectory and gaze rollouts.} Three clips showing trajectory and gaze rollouts, with and without future video.
\end{itemize}

\end{document}